\documentclass[10pt,journal,compsoc]{IEEEtran}

\ifCLASSOPTIONcompsoc
  \usepackage[nocompress]{cite}
\else
  \usepackage{cite}
\fi

\usepackage[utf8]{inputenc} 
\usepackage[T1]{fontenc}    
\usepackage{hyperref}       
\usepackage{url}            
\usepackage{booktabs}       
\usepackage{amsfonts}       
\usepackage{nicefrac}       
\usepackage{microtype}      
\usepackage{graphicx,float,url,multirow,makecell,booktabs,color,subfigure,caption,setspace}
\usepackage{amsmath}
\usepackage{cite}
\usepackage{psfrag}
\usepackage{epsfig,latexsym}
\usepackage{times}
\usepackage{threeparttable}
\usepackage{epsfig}
\usepackage{graphicx}
\usepackage{amsmath}
\usepackage{amssymb}
\usepackage{graphicx}
\usepackage{epstopdf}
\usepackage{algorithm}
\usepackage{algorithmic}

\newcommand{\tabincell}[2]{\begin{tabular}{@{}#1@{}}#2\end{tabular}}

\def\abovestrut#1{\rule[0in]{0in}{#1}\ignorespaces}
\def\belowstrut#1{\rule[-#1]{0in}{#1}\ignorespaces}
\def\abovespace{\abovestrut{0.01in}}
\def\belowspace{\belowstrut{-0.01in}}

\ifCLASSINFOpdf

\else

\fi

\begin{document}

\title{Predicting Head Movement in Panoramic Video: \\A Deep Reinforcement Learning Approach}

\author{Mai~Xu*,~\IEEEmembership{Senior Member,~IEEE,}
        Yuhang~Song*, Jianyi Wang, Minglang~Qiao, Liangyu Huo
        and~Zulin Wang
\IEEEcompsocitemizethanks{\IEEEcompsocthanksitem  M. Xu, Y. Song, J. Wang, M. Qiao, L. Huo and Z. Wang are with the School of Electronic and Information
Engineering, Beihang University, Beijing, 100191 China\protect\\
E-mail: MaiXu@buaa.edu.cn
\IEEEcompsocthanksitem M. Xu and Y. Song contribute equality to this work.
\IEEEcompsocthanksitem This work was supported by the National Nature Science Foundation of China under Grant 61573037 and by the Fok Ying Tung Education Foundation under Grant 151061.}
\thanks{Manuscript received Oct 1, 2017; revised Dec 30, 2017.}}

\markboth{Journal of \LaTeX\ Class Files,~Vol.~14, No.~8, August~2015}%
{Shell \MakeLowercase{\textit{et al.}}: Bare Advanced Demo of IEEEtran.cls for IEEE Computer Society Journals}

\IEEEtitleabstractindextext{%
\begin{abstract}
Panoramic video provides immersive and interactive experience by enabling humans to control the field of view (FoV) through head movement (HM). Thus, HM plays a key role in modeling human attention on panoramic video. This paper establishes a database collecting subjects' HM in panoramic video sequences. From this database, we find that the HM data are highly consistent across subjects. Furthermore, we find that deep reinforcement learning (DRL) can be applied to predict HM positions, via maximizing the reward of imitating human HM scanpaths through the agent's actions. Based on our findings, we propose a DRL-based HM prediction (DHP) approach with offline and online versions, called offline-DHP and online-DHP. In offline-DHP, multiple DRL workflows are run to determine potential HM positions at each panoramic frame. Then, a heat map of the potential HM positions, named the HM map, is generated as the output of offline-DHP. In online-DHP, the next HM position of one subject is estimated given the currently observed HM position, which is achieved by developing a DRL algorithm upon the learned offline-DHP model. Finally, the experiments validate that our approach is effective in both offline and online prediction of HM positions for panoramic video, and that the learned offline-DHP model can improve the performance of online-DHP.
\end{abstract}

}

\maketitle

\IEEEdisplaynontitleabstractindextext

\IEEEpeerreviewmaketitle

\ifCLASSOPTIONcompsoc
\IEEEraisesectionheading{\section{Introduction}\label{Introduction}}
\else
\section{Introduction}
\label{Introduction}
\fi

\begin{figure*}
	\begin{center}

      \subfigure[]{\includegraphics[width=0.56\columnwidth]{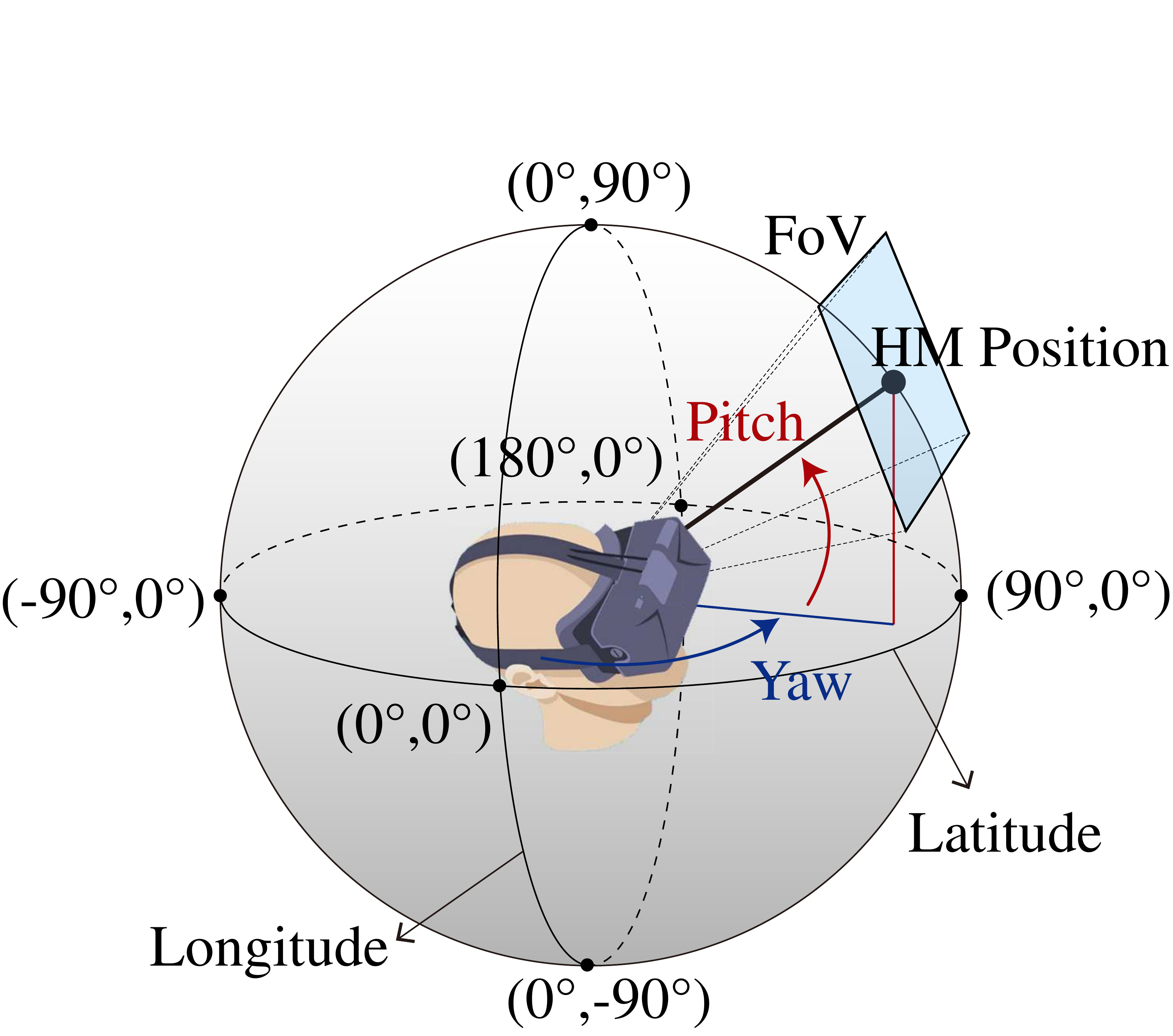}}
		\subfigure[]{\includegraphics[width=1.34\columnwidth]{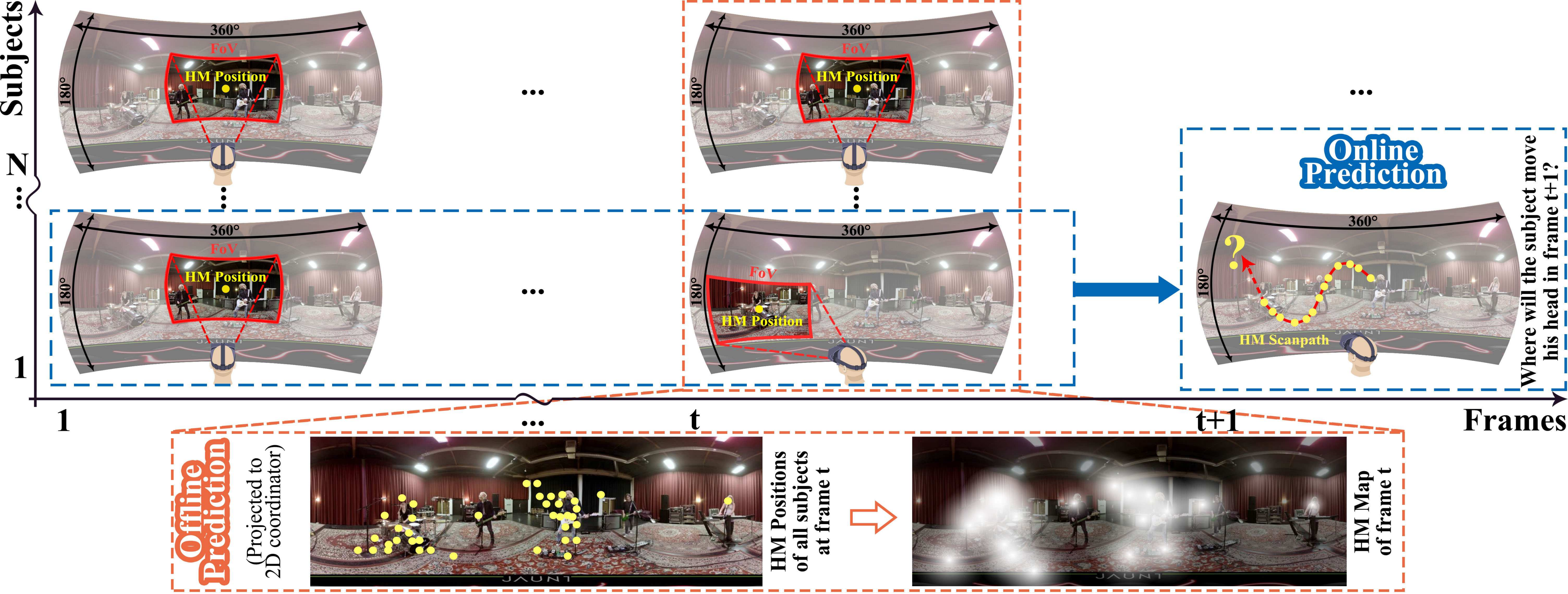}}
		\caption{\footnotesize{(a) Illustration for head movement (HM) when viewing panoramic video. (b) Demonstration for FoVs and HM positions across different subjects. The heat map of HM positions from all subjects is also shown, which is defined as the HM map.}}
		\label{fig-one}
	\end{center}
\end{figure*}

\IEEEPARstart{D}{uring} the past years, panoramic video \cite{neumann2000immersive} has become increasingly popular due to its immersive and interactive experience.
To achieve this immersive and interactive experience, humans can control the field of view (FoV)  in the range of $360^{\circ} \times 180^{\circ}$ by wearing head-mounted displays, when watching panoramic video.
In other words, humans are able to freely move their heads within a sphere to make their FoVs focus on the attractive content (see Figure \ref{fig-one} for an example).
The content outside the FoVs cannot be observed by humans, i.e., not given any attention by the viewer.
Consequently, head movement (HM) plays a key role in deploying human attention on panoramic video.
HM prediction thus emerges as an increasingly important problem in modeling attention on panoramic video.
In fact, human attention on panoramic video is composed of two parts: HM and eye fixations.
HM determines FoV as the region to be seen in panoramic video through the position of HM sampled at each frame, called the HM position in this paper. Meanwhile, eye fixations decide which region can be captured at high resolution (i.e., fovea) within the FoV.
Accordingly, HM prediction is the first step towards modeling human attention. Given the predicted HM, the eye fixations within the FoV can be further estimated using the conventional saliency detection methods \cite{borji2013state} for 2D video.
The same as traditional 2D video, the attention model can be extensively utilized in many areas of panoramic video, such as region-of-interest compression \cite{de2016video}, visual quality assessment \cite{gaddam2016tiling, Xu17}, rendering \cite{stengel2016gaze}, synopsis \cite{Pritch08}, and automatic cinematography \cite{su2016pano2vid}.

Unfortunately, few approaches have been proposed in modeling human attention on panoramic video, especially predicting the HM.
Benefiting from the most recent success of deep reinforcement learning (DRL) \cite{mnih2016asynchronous}, this paper proposes a DRL-based HM prediction (DHP) approach for modeling attention on panoramic video.
The proposed approach applies DRL rather than supervised learning. It is because DRL maximizes the accumulated \textit{reward} of the \textit{agent's} \textit{actions}, such that the predicted HM scanpaths can simulate the long-term HM behaviors of humans.
In fact, HM prediction can be classified into two categories: offline and online prediction. In this paper, the offline HM prediction is used for modeling attention of multiple subjects on panoramic video, whereas the online prediction is used to predict the next HM position of a single subject, based on the ground-truth of his/her HM positions at the current and previous frames.
In this paper, our DHP approach includes both online and offline HM prediction, named offline-DHP and online-DHP, respectively.
The codes for our offline-DHP and online-DHP approaches are downloadable from \url{https://github.com/YuhangSong/DHP}.

To our best knowledge, there exists no offline work to predict the HM positions of multiple subjects in viewing panoramic video. The closest work is saliency detection on 2D video \cite{borji2013state}. The earliest approach for saliency detection was proposed by \textit{Itti et al.} \cite{itti1998model}, in which the features of color, intensity and orientation are combined to generate the saliency map of an image. Later, \textit{Itti et al.} \cite{itti2004automatic} proposed adding two features to \cite{itti1998model}, namely, motion and flicker contrast, for video saliency detection. Recently, several advanced approaches have been proposed for video saliency prediction. These advanced works include the earth mover's distance approach \cite{lin2013visual} and the Boolean map-based saliency model (BMS) \cite{zhang2016exploiting}.
Most recently, deep learning has been successfully applied in the works of saliency detection, such as SALICON \cite{huang2015salicon} and Liu's approach \cite{Liu2017cvpr}.
Saliency detection differs from the offline prediction of HM positions in two aspects.
(1) The input to saliency detection is 2D video in a plane, whereas panoramic video is a sphere (Figure \ref{fig-one}).
Saliency detection can be applied to panoramic video that is projected from sphere to 2D plane, but projection normally causes distortion or content discontinuity, degrading the performance of predicting HM positions.
(2) More importantly, saliency detection in 2D video assumes that humans are able to view all the content of each video frame.
However, this assumption does not hold for panoramic video, as subjects can only see a limited range of the FoV at a single sight, rather than the full panoramic range of $360^{\circ} \times 180^{\circ}$.

In fact, different FoVs of panoramic video are accessible to subjects via changing the positions of HM \cite{lowe2015visualization}.
In this paper, we find that different subjects are highly consistent in terms of HM positions.
This finding is based on establishing and analyzing a new database, which consists of the HM data of 58 subjects viewing 76 panoramic video sequences.
Then, we propose the offline-DHP approach to predict the consistent HM positions on panoramic video via generating the HM map for each single frame.
The HM maps are in the form of a sphere, and the positions in the HM maps are thus represented by the longitude and latitude in the geographic coordinate system (GCS) \cite{Goodchild2007}. This paper visualizes the spherical HM maps by projecting them onto the 2D plane.
The offline prediction of Figure \ref{fig-one}-(b) demonstrates an example of the ground-truth HM map for a panoramic video frame. Similar to the saliency maps of 2D video, the HM maps of panoramic video are obtained by convoluting the HM positions with a 2D Gaussian filter\footnote{The two dimensions of the Gaussian filter are longitude and latitude, respectively.}.

Specifically, our offline-DHP approach yields the HM maps of panoramic video via predicting the HM scanpaths of multiple \textit{agents}, since subjects interactively control their HM along with some scanpaths according to video content.
First, we find from our database that the HM scanpaths of different subjects are highly consistent.
Meanwhile, subjects are normally initialized to view the center of the front region in the beginning frames of panoramic video.
Therefore, the HM positions at the subsequent frames can be yielded on the basis of the predicted scanpaths.
Additionally, we find from our database that the magnitudes and directions of HM scanpaths are similar across subjects.
In light of these findings, our offline-DHP approach models both the magnitudes and directions of HM scanpaths as the \textit{actions} of multiple DRL \textit{agents} and takes the viewed panoramic content as the \textit{observation} of the \textit{environment}.
As such, the DRL model can be learned to predict HM positions.
In training the DRL model, a \textit{reward} is designed to measure the difference of \textit{actions}  between the DRL \textit{agents} and subjects, indicating how well the \textit{agents} imitate humans in terms of HM scanpaths.
Then, the \textit{reward} is optimized to learn the parameters in the DRL model.
Given the learned model, the HM maps of panoramic video are generated upon the predicted HM positions, obtained from the scanpaths of several \textit{agents} in multiple DRL workflows.

For online HM prediction, the latest work of \cite{hu2017deep} proposed a deep 360 pilot, which automatically shifts viewing direction (equivalent to the HM position) when watching panoramic video. Specifically, the salient object is detected and tracked across panoramic video frames, via leveraging a region-based convolutional neural network (RCNN) \cite{ren2015faster} and recurrent neural network. Given the detected salient object and previous HM positions, the deep 360 pilot predicts to transit the HM position by learning a regressor. Since the deep 360 pilot relies heavily on one salient object, it is only suitable for some specific scenes that include one salient object, e.g., the sports scenes in \cite{hu2017deep}. It is still challenging to predict HM positions online for generic panoramic video, which may include more than one salient object (e.g., the panoramic video in the online prediction of Figure \ref{fig-one}-(b)). In this paper, we propose an online approach, namely online-DHP, to predict the HM positions on generic panoramic video. In contrast to \cite{hu2017deep}, our online-DHP approach does not need to detect the salient object using the RCNN. Rather, our online-DHP approach is based on attention-related content by leveraging the learned model of our offline-DHP approach. Then, a DRL algorithm is developed in our online-DHP approach to predict the HM positions in an online manner. Specifically, in the DRL algorithm, the \textit{agent} decides the \textit{action} of the HM scanpath in the next frame, according to the ground-truth of the previous HM scanpath and \textit{observation} of video content. Consequently, the HM positions at the incoming frames can be predicted for our online-DHP approach.

This paper is the first attempt to apply the DRL algorithm in modeling human attention on panoramic video. The main contributions of this paper are three-fold:
\begin{itemize}
\item We establish a new panoramic video database that consists of HM positions of 58 subjects  across 76 panoramic video sequences, with a thorough analysis of their HM data.

\item We propose an offline-DHP approach to detect HM maps of panoramic video, and this approach predicts the consistent HM positions of multiple subjects.

\item We develop an online-DHP approach to predict the HM position of one subject at the next frame, based on the video content and HM scanpath till the current frame.

\end{itemize}

\section{Related work}

\subsection{Saliency detection}
The only approach for predicting the HM positions of panoramic video is the most recent work of \cite{su2016pano2vid}, in which Pano2Vid was proposed to obtain the FoV at each panoramic video frame.
However, Pano2Vid primarily focuses on virtually generating a potential HM position at one frame, rather than modeling HM maps of multiple subjects at this frame.
The closest work on predicting HM maps is saliency detection for 2D video, which is briefly reviewed in the following.

Saliency detection aims to predict the visual attention of humans on 2D video, by generating saliency maps of video frames. The studies on visual saliency began in 1998, when Itti and Koch \cite{itti1998model} found that the features of intensity, color and orientation in an image can be employed to detect its saliency map. Subsequently, they extended their work to video saliency detection \cite{itti2004automatic}, in which two dynamic features of motion and flicker contrast are combined with \cite{itti1998model} to detect saliency in 2D video. Both \cite{itti1998model} and \cite{itti2004automatic} are heuristic approaches for detecting saliency, since they utilize the understanding of the human vision system (HVS) to develop the computational models. Recently, some advanced heuristic approaches, e.g.,  \cite{itti2009bayesian, boccignone2008nonparametric, zhang2009sunday, guo2010novel, ren2013regularized, lin2013visual, zhang2016exploiting, hossein2015many, xu2017learning}, have been proposed to detect saliency in 2D video.
Specifically, \cite{itti2009bayesian} proposed a novel feature called \textit{surprise}, which measures how the visual change attracts human observers, based on the Kullback-Leibler divergence between spatio-temporal posterior and prior beliefs.
Given the feature of \textit{surprise}, a Bayesian framework was developed in \cite{itti2009bayesian} for video saliency detection.
Some other Bayesian frameworks \cite{boccignone2008nonparametric, zhang2009sunday}  were also developed to detect video saliency.
Besides, Lin \textit{et al.} \cite{lin2013visual} quantified the earth mover's distance to measure the center-surround difference in spatio-temporal receptive field, generating saliency maps for 2D video.
Zhang \textit{et al.} \cite{zhang2016exploiting} explored the surround cue for saliency detection, by characterizing a set of binary images with random thresholds on color channels.
Recently, \cite{hossein2015many} and \cite{xu2017learning} have investigated that some features (e.g., motion vector) in compressed domain are of high correlation with human attention, and these features are thus explored in video saliency detection.

Benefiting from the most recent success of deep learning, deep
neural networks (DNNs) \cite{Vig_2014_CVPR, huang2015salicon, kruthiventi2015deepfix, Liu_2015_CVPR, wang2016RCNN, bazzani2016recurrent, SalGAN_2017,  Liu2017cvpr, bak2016two,wang2017deep, Kummerer_2017_ICCV, Palazzi_2017_ICCV} have also been developed to detect 2D video saliency, rather than
exploring the HVS-related features as in heuristic saliency detection
approaches. These DNNs can be viewed as data-driven approaches.
For static saliency detection, SALICON \cite{huang2015salicon} fine-tuned the existing
convolutional neural networks (CNNs), with a new saliency-related
loss function. In \cite{Liu_2015_CVPR}, the architecture of multi-resolution CNN was developed for detecting saliency of images.
In \cite{Kummerer_2017_ICCV}, a readout architecture was proposed to predict human attention on static images, in which both DNN features and low-level (isotropic contrast) features are considered.
For dynamic saliency detection, \cite{bazzani2016recurrent}  leveraged a
deep convolutional 3D network to learn the representations
of human attention on 16 consecutive frames, and then a long short-term
memory (LSTM) network connected with a mixture density
network was learned to generate saliency maps using Gaussian
mixture distribution. Similarly, Liu et al. \cite{Liu2017cvpr} combined a CNN
and multi-stream LSTM to detect saliency in video with multiple
faces. Moreover, other DNN structures have been developed to
detect either static saliency \cite{Vig_2014_CVPR, kruthiventi2015deepfix, wang2016RCNN, SalGAN_2017} or dynamic saliency \cite{bak2016two,bazzani2016recurrent,wang2017deep, Palazzi_2017_ICCV}.

Although saliency detection has been thoroughly studied in predicting eye movement in 2D video, no work has been developed to predict HM positions on panoramic video.
Similar to saliency detection for 2D video, this paper proposes generating HM maps that represent the HM positions of multiple subjects.
To obtain the HM maps of panoramic video, the HM positions are predicted by estimating the HM scanpaths of several \textit{agents}. Similarly, in the saliency detection area, there exist some works \cite{Wang2011, Sun12, Liu_2013_ICCV, Jiang2016, Assens_2017_ICCV, Shao2017} that predict eye movement scanpaths for static images.
In \cite{Wang2011}, a computational model was developed to simulate the scanpaths of eye movement in natural images. The proposed model embeds three factors to guide eye movement sequentially, including reference sensory responses, fovea periphery resolution discrepancy, and visual working memory. Sun \textit{et al.} \cite{Sun12} proposed modeling both saccadic scanpaths and visual saliency of images, on the basis of super Gaussian component (SGC) analysis. Recently, data-driven approaches have been proposed to learn the scanpaths of eye movement in static images, such as the hidden Markov model in \cite{Liu_2013_ICCV} and least-squares policy iteration (LSPI) in \cite{Jiang2016}.
Most recently, deep learning has been utilized in \cite{Assens_2017_ICCV, Shao2017} for predicting the eye movement scanpaths in static images.
However, to our best knowledge, there is no work on predicting the HM scanpaths on panoramic video.

In this paper, a DRL approach is developed for predicting the actions of the HM scanpaths from multiple \textit{agents}.
The actions are decided based on the environment of the panoramic video content, the features of which are automatically learned and then extracted by a DNN. Thus, our approach takes advantage
of both deep learning and reinforcement learning, driven by the HM data of our panoramic video database. Note that although few works apply DRL to predict human attention, the attention model is widely used in the opposite direction to improve the performance of reinforcement learning, e.g., \cite{minut2001reinforcement, mnih2014recurrent, jaderberg2016reinforcement, wang2016dueling}.

\subsection{Virtual cinematography}
Virtual cinematography of panoramic video, which directs an imaginary camera to virtually capture natural FOV, was proposed in \cite{foote2000flycam, sun2005region, su2016pano2vid, hu2017deep, lin2017tell}. In general, virtual cinematography attempts to agree with the HM positions of humans at each panoramic video frame. The early work of \cite{foote2000flycam} proposed cropping the object-of-interest in panoramic video, such that the natural FOV can be generated for virtual cinematography. Later, in \cite{sun2005region}, the cropped object-of-interest is tracked across frames by a Kalman filter, for automatically controlling the virtual camera in virtual cinematography of panoramic video. The approach of \cite{sun2005region} can work on both compressed and uncompressed domains, because two methods were developed to detect the object-of-interest in compressed and uncompressed domains. The works of \cite{foote2000flycam, sun2005region} were both designed for the task of online virtual cinematography. These works can be considered as heuristic approaches, which are not trained or even evaluated on the ground-truth HM data of human subjects.

Most recently, data-driven approaches have boosted the development of virtual cinematography for panoramic video. Specifically, Pano2Vid \cite{su2016pano2vid} learns to generate natural FOV at each panoramic frame. However, the learning mechanism of Pano2Vid is offline. In fact, natural FOV can be estimated at each frame in an online manner, which uses the observed HM positions of the previous frames to correct the estimation of natural FOV at the current frame. To this end, online virtual cinematography \cite{hu2017deep, lin2017tell} has been studied in a data-driven way.
Specifically, a state-of-the-art virtual cinematography  approach, the deep 360 pilot,  was proposed in \cite{hu2017deep}, which is a deep-learning-based \textit{agent} that smoothly tracks the object-of-interest for panoramic video. In other words, the \textit{agent} transits the HM position across video frames to track the key object detected by the RCNN, given the observed HM positions at previous frames. Consequently, natural FOV can be generated online for automatically displaying the object-of-interest in virtual cinematography of panoramic video. In fact, object-of-interest tracking in panoramic video refers to continuously focusing and refocusing the intended targets. Both focusing and refocusing require a subject to catch up the object. Such a task is challenging in extreme-sports video, as the object-of-interest may be moving fast. Therefore, Lin \textit{et al.} \cite{lin2017tell} investigated two focus assistance techniques to help the subject track the key object in viewing panoramic video, in which the potential HM position attended to the object-of-interest needs to be determined and provided for the subject.

The above approaches of \cite{foote2000flycam, sun2005region, su2016pano2vid, hu2017deep, lin2017tell}  all depend on the detector of the object-of-interest. Thus, they can only be applied in some specific panoramic video with salient objects, such as video conferencing or classroom scenes in \cite{foote2000flycam, sun2005region} and the sports video in \cite{su2016pano2vid, hu2017deep, lin2017tell}. Different from these conventional approaches, our online-DHP approach is based on the learned model of our offline approach, which encodes HM-related content rather than detecting the object-of-interest. Consequently, our approach is object free, thus more suitable for generic panoramic video.

\section{Database establishment and findings}
\label{Database_establishment_and_analysis}

In this section, we collect a new database that includes 76 panoramic video sequences with the HM data of 58 subjects, called the PVS-HM database.
Along with the HM data, the eye fixation data of 58 subjects are also obtained in our PVS-HM database.
Our PVS-HM database allows quantitative analysis of subjects' HM on panoramic video, and it can also be used for learning to predict where humans look at panoramic video. Our database is available at  \url{https://github.com/YuhangSong/dhp} for facilitating future research. In the following, we present how we conducted the experiment to obtain the PVS-HM database.

First, we selected 76 panoramic video sequences from YouTube and VRCun, with resolutions ranging from 3K to 8K.
As shown in Table 1 of the supplemental material, the content of these sequences is diverse, including computer animation, driving, action sports, movies, video games, scenery, and so forth.
Then, the duration of each sequence was cut to be from 10 to 80 seconds (averagely 26.9 seconds), such that fatigue can be reduced when viewing panoramic video.
To ensure video quality, all panoramic video sequences were compressed using H.265 \cite{Sullivan2013Overview} without any change in bit-rates.
Note that the audio tracks were removed to avoid the impact of acoustic information on visual attention.

In our experiment, 58 subjects (41 males and 17 females, ranging in age from 18 to 36) wore the head-mounted display of an HTC Vive to view all 76 panoramic video sequences at a random display order.
When watching panoramic video, the subjects were seated on a swivel chair and were allowed to turn around freely, such that all panoramic regions are accessible.
To avoid eye fatigue and motion sickness, the subjects had a 5-minute rest after viewing each session of 19 sequences.
With the support of the software development kit of the HTC Vive, we recorded the posture data of each subject as they viewed the panoramic video sequences.
Based on the recorded posture data, the HM data of all 58 subjects at each frame of the panoramic video sequences were obtained and stored for our PVS-HM database, in terms of longitude and latitude in the GCS.
In addition to the recorded HM data, the eye fixations were also captured by the VR eye-tracking module aGlass\footnote{When subjects viewing panoramic video, the aGlass device is able to capture the eye fixations within FoV at less than $0.5^{\circ}$ error. See \url{http://www.aglass.com/?lang=en} for more details about this device.}, which was embedded in the head-mounted display of the HTC vive.

Then, we mine our PVS-HM database to analyze the HM data of different subjects across panoramic video sequences.
Specifically, we have the following five findings, the analysis of which is presented in the supplemental material.
1) The HM positions on panoramic video possess front center bias (FCB).
2) When watching panoramic video, different subjects are highly consistent in HM positions.
3) The magnitude of HM scanpaths is similar across subjects, when viewing the same regions of panoramic video.
4) The direction of HM scanpaths on panoramic video is highly consistent across subjects.
5) Almost $50\%$ subjects are consistent in one HM scanpath direction (among 8 uniformly quantized directions), and over $85\%$ of subjects are consistent in three directions for HM scanpaths.

\begin{figure*}
	\begin{center}
        \vspace{-1em}
		\centerline{\includegraphics[width=2.0\columnwidth]{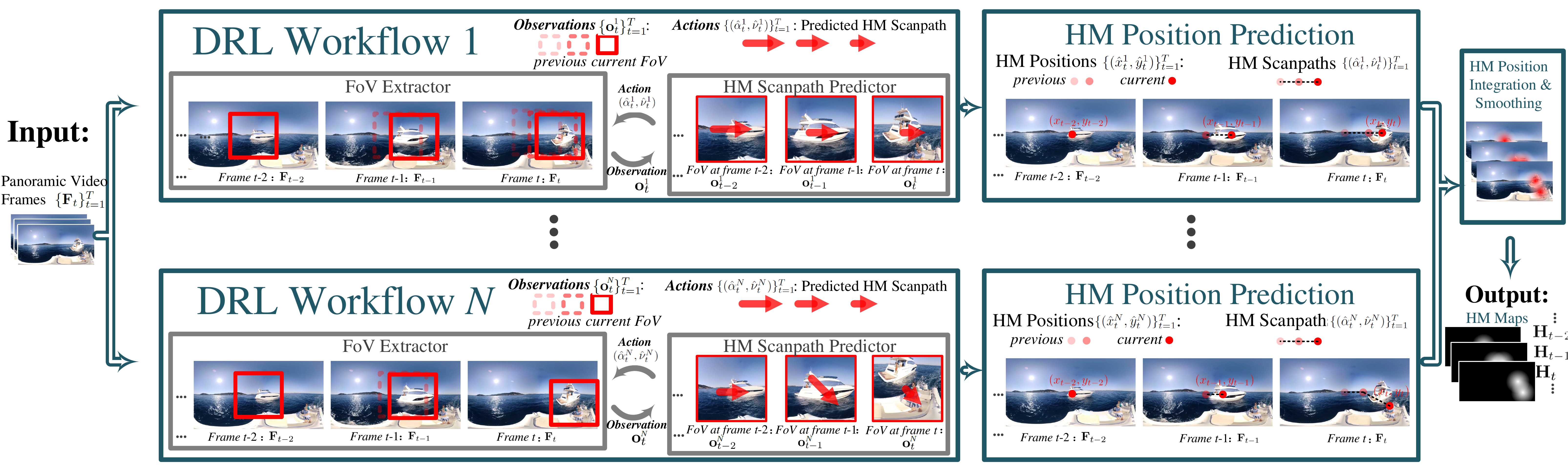}}
        \vspace{-1em}
		\caption{\footnotesize{Overall framework of the offline-DHP approach.}}
		\label{main-framework}
	\end{center}
\vspace{-1em}
\end{figure*}

\section{Offline-DHP approach}\label{sec::offline-DHP}

\subsection{Framework of offline-DHP}
\label{framework}

In this section, we present our offline-DHP approach, in light of our findings in Section \ref{Database_establishment_and_analysis}.
Figure \ref{main-framework} shows the overall framework of our approach, in which the multiple DRL workflows are embedded to generate the HM maps of input panoramic video frames.
The procedure and notations of Figure \ref{main-framework} are presented in the following.

As shown in Figure \ref{main-framework}, the input to our offline-DHP approach is the panoramic video frames $\{\mathbf{F}_t\}_{t=1}^{T}$ with frame number $t$ ranging from $1$ to $T$.
Since \textit{Finding 2} has shown that the HM positions are highly consistent across different subjects,  we propose to generate the HM maps $\{\mathbf{H}_t\}_{t=1}^{T}$ for modeling human attention on panoramic video, viewed as the output of our offline-DHP approach. The HM map $\mathbf{H}_t$ of frame $t$ represents the probability of each pixel being the HM position.
Assuming that $\{(\hat{x}^n_t, \hat{y}^n_t)\}_{n=1}^{N}$ are the HM positions at the $t$-th frame, $\mathbf{H}_t$ is obtained by convoluting $\{(\hat{x}^n_t, \hat{y}^n_t)\}_{n=1}^{N}$ with a 2D Gaussian filter, similar to the saliency maps of 2D video.
Here, $n$ means the $n$-th HM position and $N$ is the total number of HM positions.

Because \textit{Finding 5} has indicated that the HM scanpaths of different subjects are consistent in more than one direction, the HM positions $\{({x}^m_t, {y}^m_t)\}_{m=1}^{M}$ of subjects $\{m\}_{m=1}^{M}$ may be different from each other. Accordingly, this paper assumes that the number of predicted HM positions $N$ is equivalent to $M$ at each frame, for predicting the HM positions of all subjects.
In other words, to obtain $(\hat{x}^n_t, \hat{y}^n_t)$, our offline-DHP approach applies one DRL workflow to estimate the HM positions of one subject.
Then, $N$ DRL workflows are run to obtain $N$ HM positions $\{(\hat{x}^n_t, \hat{y}^n_t)\}_{n=1}^{N}$ at frame $t$, simulating the ground-truth HM positions of $M$ ($=N$) subjects at this frame.
At a panoramic frame, each of the DRL workflows works independently to generate an HM position by randomly sampling actions based on a learned \textit{policy} $\pi_t$, which is modeled as the predicted probability distribution of the HM direction at frame $t$.
Note that all DRL workflows share the same \textit{policy} $\pi_t$ in our approach.

Let $\hat{\alpha}^n_t$ and $\hat{\nu}^n_t$ be the direction and magnitude of the predicted HM scanpath at frame $t$, obtained from the $n$-th DRL workflow.
They are both viewed as the $actions$ of the DRL workflow.
In a single DRL workflow, $\{(\hat{x}^n_t, \hat{y}^n_t)\}_{t=1}^T$ can be modeled by determining a series of \textit{actions}: $\{\hat{\alpha}^n_t\}_{t=1}^T$ and $\{\hat{\nu}^n_t\}_{t=1}^T$.
It is worth pointing out that $\{\hat{\alpha}^n_t\}_{t=1}^{T}$ and $\{\hat{\nu}^n_t\}_{t=1}^{T}$ are predictable as the \textit{actions} of the DRL workflow, since \textit{Findings 3} and \textit{4} have indicated that subjects are consistent in the magnitudes and directions of HM scanpaths.
The direction and magnitude of the ground-truth HM scanpath are denoted by $\alpha^m_t$ and $\nu^m_t$ for the $m$-th subject at frame $t$.

As can be seen in Figure \ref{main-framework}, in each workflow, one HM scanpath is generated through the interaction between the FoV extractor\footnote{Note that the extracted FoV is $103^{\circ} \times 60^{\circ}$, which is the same as the setting of the head-mounted display.} and HM scanpath predictor.
Specifically, $\{\mathbf{o}^n_t\}_{t=1}^{T}$ denotes the FoVs of frames from $1$ to $T$ in the $n$-th DRL workflow. Figure \ref{main-framework} shows that FoV $\mathbf{o}^n_t$ is extracted via making its center locate at the HM position $(\hat{x}^n_t,\hat{y}^n_t)$, in which $(\hat{x}^n_t,\hat{y}^n_t)$ is generated by the predicted \textit{action} of HM scanpath $(\hat{\alpha}^n_{t-1},\hat{\nu}^n_{t-1})$ at the previous video frame.
Then, the content of the extracted FoV works as the \textit{observation} of DRL, for predicting the next \textit{action} of HM scanpath $(\hat{\alpha}^n_{t},\hat{\nu}^n_{t})$.
The HM scanpath generated by each DRL workflow is forwarded to obtain HM positions at incoming frames.
Subsequently, the HM positions from multiple DRL workflows are integrated, and then smoothed by a 2D Gaussian filter.
Finally, the HM maps $\{\mathbf{H}_t\}_{t=1}^{T}$ of the panoramic video are obtained, which model the heat maps for the HM positions at each frame.

\subsection{DRL model of the offline-DHP approach}
\label{train}
\begin{figure*}
	\begin{center}
		\centerline{\includegraphics[width=1.8\columnwidth]{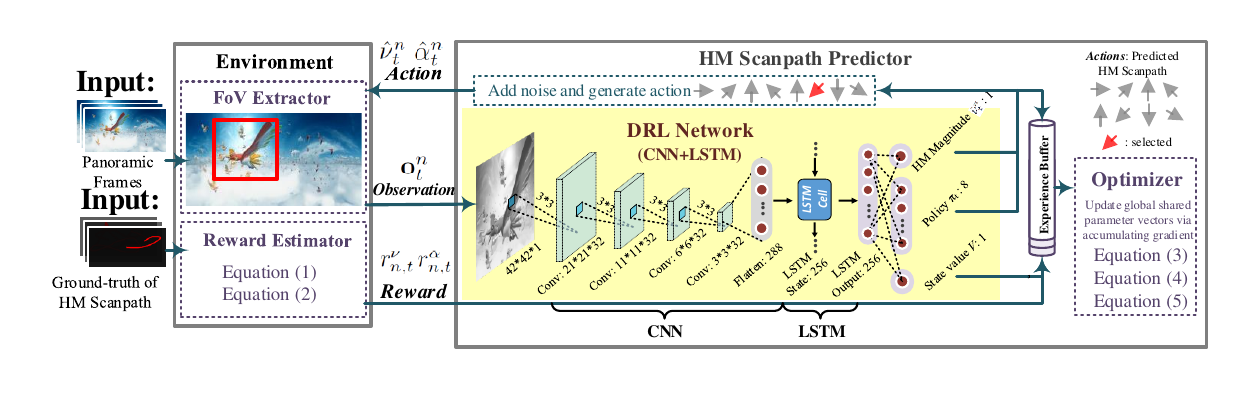}}
		\caption{\footnotesize{Framework of training the DRL model to obtain each DRL workflow of the offline-DHP approach (Figure \ref{main-framework}).}}
		\label{train-framework}
	\end{center}
\end{figure*}

As described in Section \ref{framework}, the DRL workflow is a key component in our offline-DHP framework, which targets at predicting the HM scanpaths.
This section presents how to train the DRL model of each workflow for predicting the HM maps.
Note that our offline-DHP approach runs multiple workflows to train one global-shared model, the same as the asynchronous DRL method \cite{mnih2016asynchronous}.
In this section, we take the $n$-th workflow as an example.
Figure \ref{train-framework} shows the framework of training the DRL model.
As shown in this figure, the FoV of the input video frame is extracted based on the \textit{action} of the HM scanpath predicted at the previous frame.
The extracted FoV, as the \textit{observation}, is then fed into the DRL network.
The structure of the DRL network follows \cite{mnih2016asynchronous}, which has four 32-filter convolutional layers (size: $21\times 21$, $11\times 11$, $6\times 6$ and $3\times 3$), one flatten layer (size: $288$) and LSTM cells (size: 256).
Then, the 256-dimensional LSTM feature $\mathbf{f}^n_{t}$ is output at frame $t$, as part of the \textit{observed state} in the $n$-th DRL workflow.
In addition, the \textit{reward}, which measures the similarity between the predicted and ground-truth HM scanpaths, is estimated to evaluate the \textit{action} made by the DRL model. Then, the \textit{reward} is used to make decision on the \textit{action} through the DRL model, i.e., the HM scanpath at the current frame.
In this paper, we denote  $r^{\alpha}_{n,t}$ and $r^{\nu}_{n,t}$ as the  \textit{rewards} for evaluating \textit{actions} $\hat{\alpha}^n_t$ and $\hat{\nu}^n_t$, respectively, in the $n$-th DRL workflow.
Finally, the \textit{environment} of our DRL model is comprised by the \textit{observation} of the extracted FoV and the \textit{reward} of HM scanpath prediction.

In training the DRL model,  the \textit{environment} interacts with the HM scanpath predictor.
The interaction is achieved in our DRL model through the following procedure.\\
(1) At frame $t$, the FoV extractor obtains the current $\textit{observation}$ $\mathbf{o}^n_t$ ($103^{\circ} \times 60^{\circ}$) from the input video frame $\mathbf{F}_t$, according to the predicted HM position $(\hat{x}^n_t,\hat{y}^n_t)$.
In our work, $\mathbf{o}^n_{t}$ is projected onto the 2D region and is then down-sampled to $42\times42$.\\
(2) The current $\mathbf{o}^n_t$ and the LSTM feature $\mathbf{f}^n_{t-1}$ from  the last frame are delivered to the DRL network in the HM scanpath predictor.
In our work, the DRL network contains four convolutional layers and one LSTM layer \cite{hausknecht2015deep}, which are used to extract the spatial and temporal features, respectively. The details about the architecture of the DRL network can be found in Figure \ref{train-framework}.\\
(3) At frame $t$, the DRL network produces the LSTM feature $\mathbf{f}^n_{t}$, HM scanpath magnitude $\hat{\nu}^n_{t}$ and policy $\pi_{t}$. Here, $\pi_{t}$ is modeled by the probability distribution over the \textit{actions} of HM scanpath directions.\\
(4) Given $\pi_{t}$, the HM scanpath predictor randomly samples an \textit{action} $\hat{\alpha}^n_t$ with standard deviation $\varepsilon$, such that the exploration is ensured in decision making. Here, $\hat{\alpha}^n_t$ includes 8 discrete directions in GCS: $\{ 0^{\circ}, 45^{\circ}, \cdots, 315^{\circ} \}$.\\
(5) \textit{Environment} is updated using $\hat{\nu}^n_t$ and $\hat{\alpha}^n_t$, leading to $(\hat{x}^n_t, \hat{y}^n_t)\longrightarrow (\hat{x}^n_{t+1},\hat{y}^n_{t+1})$. The FoV extractor returns a new \textit{observation} $\mathbf{o}^n_{t+1}$ according to the HM position $(\hat{x}^n_{t+1},\hat{y}^n_{t+1})$. The \textit{reward} estimator returns the \textit{rewards} $r^{\nu}_{n,t}$ and $r^{\alpha}_{n,t}$ in predicting $\hat{\nu}^n_t$ and $\hat{\alpha}^n_t$, based on the ground-truth HM scanpaths of $\{\nu^m_t\}_{m=1}^{M}$ and $\{\alpha^m_t\}_{m=1}^{M}$. \\
(6) A set of experiences $\{ \mathbf{o}^n_{t}, \! \mathbf{f}^n_{t-1},\! \hat{\nu}^n_t,\! \hat{\alpha}^n_t,\! r^{\nu}_{n,t},\! r^{\alpha}_{n,t} \}$ are stored in an experience buffer for frame $t$.
In addition, $\mathbf{o}^n_{t+1}$ and $\mathbf{f}^n_{t}$ are preserved for processing frame $t+1$.\\
(7) Once $t$ meets the termination condition of exceeding the maximum frame number $T$, all experiences in the buffer are delivered to the optimizer for updating the DRL network.

\textbf{Reward Estimation.}
Next, we focus on modeling the \textit{rewards} $r^{\alpha}_{n,t}$ and $r^{\nu}_{n,t}$ in determining the \textit{actions} of HM scanpaths.
When training the DRL model, our goal is to make the prediction of $\hat{\alpha}^n_{t}$ and $\hat{\nu}^n_{t}$ approach the ground-truth HM scanpaths.
Thus, the  \textit{rewards} $r^{\alpha}_{n,t}$ and $r^{\nu}_{n,t}$ can be represented by the differences from $\hat{\alpha}^n_{t}$  to $\{{\alpha}^m_{t}\}_{m=1}^M$ and from $\hat{\nu}^n_{t}$ to $\{{\nu}^m_{t}\}_{m=1}^M$, respectively.
In our approach, these differences are measured by Gaussian distributions.
We further consider the distances from predicted HM position $(\hat{x}^n_t,\hat{y}^n_t)$ to $\{(x^{m}_{t},y^{m}_{t})\}_{m=1}^M$ in calculating the \textit{rewards} of $r^{\alpha}_{n,t}$ and $r^{\nu}_{n,t}$, which are also modeled by the 2D Gaussian distribution.
This consideration is because only the consistent HM regions have similar HM scanpaths, according to the analysis of \textit{Finding 4}.
Then, $r^{\alpha}_{n,t}$ can be written as
\begin{equation}
\label{reward-alpha}
r^{\alpha}_{n,t} = \frac{1}{N}\sum_{m=1}^{M} e^{-\frac{1}{2}\left(\frac{D_d(\hat{\alpha}^n_{t}, \alpha^m_{t})}{\rho}\right)^2} e^{-\frac{1}{2}\left(\frac{D_s((\hat{x}^n_{t},\hat{y}^n_{t}),(x^m_{t},y^m_{t}))}{\varrho}\right)^2}.
\end{equation}
In \eqref{reward-alpha}, $D_d$ defines the phase difference, and $D_s$ denotes the \textit{great-circle distance} \cite{shumaker1984astronomical}. Moreover, $\rho$ and $\varrho$ are the standard deviations of Gaussian distributions, as the hyper-parameters.
In \eqref{reward-alpha},  the similarity score of $e^{-\frac{1}{2}\left(\frac{D_d(\hat{\alpha}^n_{t}, \alpha^m_{t})}{\rho}\right)^2}$ measures the similarity of HM direction between the ground-truth and \textit{agent action},
while $e^{-\frac{1}{2}\left(\frac{D_s((\hat{x}^n_{t},\hat{y}^n_{t}),(x^m_{t},y^m_{t}))}{\varrho}\right)^2}$ qualifies the validity of the corresponding similarity score in calculating the reward.

Then, given the HM direction, we can estimate its corresponding magnitude through reward  $r^{\nu}_{n,t}$. Similar to \eqref{reward-alpha}, we have
\begin{equation}
\label{reward-nu}
r^{\nu}_{n,t} \!\!=\!\! \frac{1}{N}\!\!\!\sum_{m=1}^{M}\!\!
e^{\!-\frac{1}{2}!\left(\!{\frac{\hat{\nu}^n_{t}-\nu^{m}_{t}}{\varsigma}}\!\right)\!^2} \!\! e^{\!-\frac{1}{2}\!\left(\!\frac{D_d(\hat{\alpha}^n_{t}, \alpha^m_{t})}{\rho}\!\right)\!^2} \!\!e^{\!-\frac{1}{2}\!\left(\!\frac{D_s((\hat{x}^n_{t},\hat{y}^n_{t}),(x^m_{t},y^m_{t}))}{\varrho}\!\right)\!^2},
\end{equation}
where $\varsigma$ is the hyper-parameter for the standard deviation of the HM scanpath magnitude.
As defined in \eqref{reward-nu}, $e^{\!-\frac{1}{2}\left(\!{\frac{\hat{\nu}^n_{t}-\nu^{m}_{t}}{\varsigma}}\!\right)^2}$ is the similarity score of the HM scanpath magnitude.
Reward $r^{\nu}_{n,t}$ is valid in predicting the magnitude, only if both the predicted HM position and direction are similar to the ground-truth.
Thus, $e^{\!-\frac{1}{2}\left(\!\frac{D_d(\hat{\alpha}^n_{t}, \alpha^m_{t})}{\rho}\!\right)^2}$ and $e^{\!-\frac{1}{2}\left(\!\frac{D_s((\hat{x}^n_{t},\hat{y}^n_{t}),(x^m_{t},y^m_{t}))}{\varrho}\!\right)^2}$ are introduced in \eqref{reward-nu} to determine the validity of the similarity score.

\textbf{Optimization.}
Next, we need to optimize the \textit{rewards} $r^{\alpha}_{n,t}$ and $r^{\nu}_{n,t}$, when learning the network parameters of our DRL model in Figure \ref{train-framework}.
Our offline-DHP approach applies the asynchronous DRL method \cite{mnih2016asynchronous} to learn the DRL parameters with optimized \textit{rewards}.
Hence, multiple workflows are run to interact with multiple \textit{environments} with workflow-specific parameter vectors $\{ \theta^{n}_{\nu}, \theta^{n}_{\pi}, \theta^{n}_{V} \}$, producing $\hat{\nu}^n_t$, $\hat{\pi}^n_t$ and $V$.
Here, $V$ denotes the \textit{state value} output by the DRL network, which is obtained using the same way as \cite{mnih2016asynchronous}.
Meanwhile, global-shared parameter vectors $\{ \theta_{\nu}, \theta_{\pi}, \theta_{V} \}$\footnote{As can be seen in Figure \ref{train-framework}, $\{ \theta_{\nu}, \theta_{\pi}, \theta_{V} \}$ share all CNN and LSTM layers in our offline-DHP approach, but they are separated at the output layer.} are updated via an accumulating gradient.
For more details about the workflow-specific and global-shared parameter vectors, refer to \cite{mnih2016asynchronous}.
In our approach, \textit{reward} $r^{\nu}_{n,t}$ is optimized to train $\theta_{\nu}$ as follows:
\begin{equation}
\label{opt-1}
d \theta_{\nu} \leftarrow d \theta_{\nu} + \nabla_{\theta_{\nu}^{n}} \sum_{t=1}^{T} r^{\nu}_{n,t}.
\end{equation}
Moreover, we can optimize \textit{reward} $r^{\alpha}_{n,t}$ by
\begin{equation}
\label{opt-2}
\small d \theta_{V} \leftarrow d \theta_{V} + \nabla_{\theta_{V}^{n}} \sum_{t = 1}^{T} (\sum_{i=t}^{T} \gamma^{i-t} r^{\alpha}_{n,i} - V(\mathbf{o}^n_{t}, \mathbf{f}^n_{t-1} ; \theta_{V}^{n}))^2,
\end{equation}
\begin{small}
\begin{eqnarray}
\label{opt-3}
\hspace{0.1cm} \nonumber d \theta_{\pi} \leftarrow d \theta_{\pi} +\nabla_{\theta_{\pi}^{n}} \sum_{t = 1}^{T} \log \pi( \hat{\alpha}^n_{t} | \mathbf{o}^n_{t}, \mathbf{f}^n_{t-1} ; \theta_{\pi}^{n})\cdot \\
(\sum_{i = t}^{T} \gamma^{i-t} r^{\alpha}_{n,i} - V(\mathbf{o}^n_{t}, \mathbf{f}^n_{t-1} ; \theta_{V}^{n})),
\end{eqnarray}
\end{small}
where $\gamma$ is the discount factor of Q-learning \cite{watkins1992q}.
In addition, $V(\mathbf{o}^n_{t}, \mathbf{f}^n_{t-1} ; \theta_{V}^{n})$ denotes state value $V$ obtained by $\mathbf{o}^n_{t}, \mathbf{f}^n_{t-1}$ and $\theta_{V}^{n}$; $\pi( \hat{\alpha}^n_{t} | \mathbf{o}^n_{t}, \mathbf{f}^n_{t-1} ; \theta_{\pi}^{n})$ stands for the probability of \textit{action} $\hat{\alpha}^n_{t}$ that is made by policy $\pi_t$ from $\mathbf{o}^n_{t}, \mathbf{f}^n_{t-1}$ and $\theta_{\pi}^{n}$.
Finally, based on the above equations, RMSProp \cite{tieleman2012lecture} is applied to optimize \textit{rewards} in the training data. Consequently, the workflow-specific and global-shared parameter vectors can be learned to predict HM scanpaths.
Finally, these learned parameter vectors can be used to determine the scanpaths and positions of HM through each DRL workflow in our offline-DHP approach.

\section{Online-DHP approach}\label{sec:online-approach}

\begin{figure}
\vspace{-2em}
	\begin{center}
		\centerline{\includegraphics[width=1.05\columnwidth]{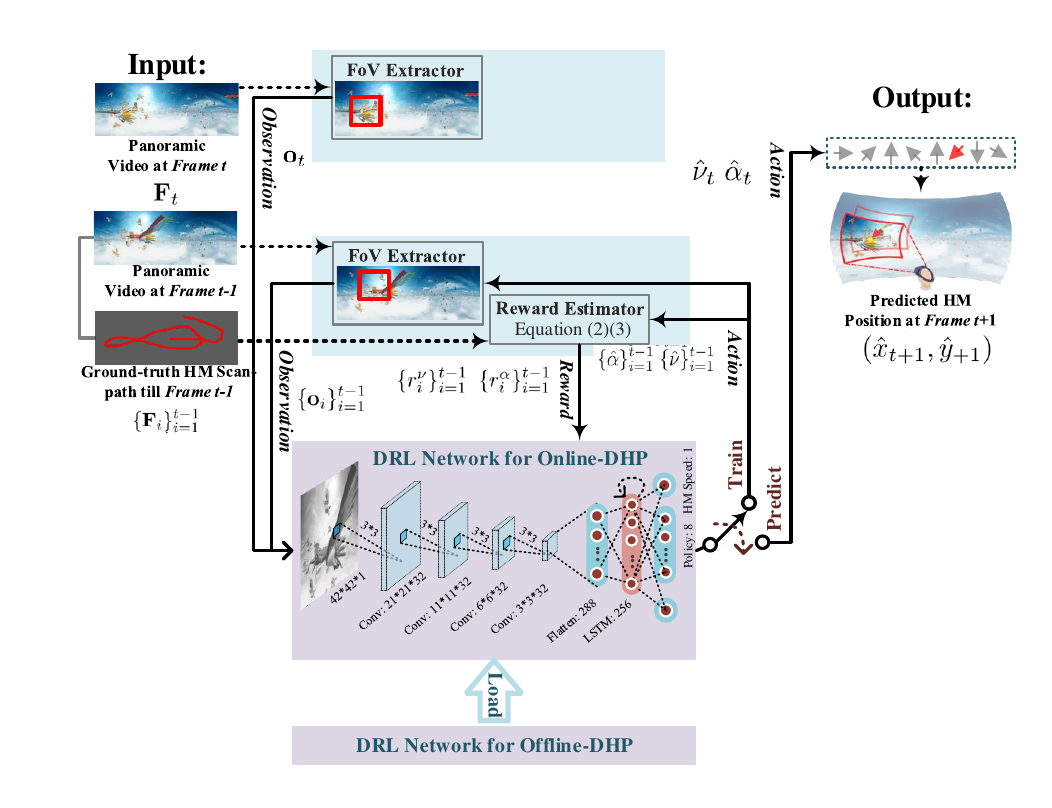}}
       \vspace{-1em}
		\caption{\footnotesize{Framework of the online-DHP approach.}}
		\label{online-framework}
	\end{center}
\vspace{-2em}
\end{figure}

In this section, we present our online-DHP approach.
The online-DHP approach refers to predicting a specific subject's HM position $(\hat{x}_{t+1},\hat{y}_{t+1})$ at frame $t+1$, given his/her HM positions $\{(x_{1},y_{1}),\ldots, (x_{t},y_{t})\}$ till frame $t$.
Note that the definitions of the notations in this section are similar to those in Section \ref{sec::offline-DHP}, and the only difference is that $n$ and $m$ are removed in all notations because there is only one subject/workflow in online-DHP.
Additionally, we define the subject as the \textit{viewer}, whose HM positions need to be predicted online.
Figure \ref{online-framework} shows the framework of our online-DHP approach.
It is intuitive that the current HM position is correlated with the previous HM scanpaths and video content.
Therefore, the input to our online-DHP framework is the \textit{viewer's} HM scanpath $\{(\alpha_1,\nu_1),\ldots, (\alpha_{t-1},\nu_{t-1})\}$ and frame content $\{\mathbf{F}_1, \ldots, \mathbf{F}_t \}$, and the output is the predicted HM position $(\hat{x}_{t+1},\hat{y}_{t+1})$ at the next frame for the \textit{viewer}.
This can be viewed as online prediction of HM positions $\{(\hat{x}_{t},\hat{y}_{t})\}_{t=1}^{T}$ . To this end, our online-DHP consists of two stages: the training and prediction stages.
In the first stage, the parameters of the DRL network are trained.
In the second stage, the \textit{action} of the HM scanpath is generated from the trained DRL network, to predict the HM position online.
In the following, we discuss these two stages in more detail.

\subsection{Stage I: Training}
At the beginning frame, the HM position $(\hat{x}_1,\hat{y}_1)$ of the \textit{viewer} is initialized to be the center of the front region, which is the general setting of  the panoramic video player. Then, the trained DRL network of offline-DHP is loaded as the initial DRL network for online prediction, both sharing the same structure.
The reason for loading the offline-DHP network is that it encodes the knowledge of HM-related features.
Later, this initial DRL network is fine-tuned by the \textit{viewer's} HM scanpath at incoming frames.

Next, we focus on the algorithm for training the DRL network in our online-DHP approach. As previously mentioned, the initial parameters of the DRL network at the first frame are directly from those of offline-DHP. At each of the incoming frames, several episodes are run to update the DRL network for online-DHP. The following summarizes the procedure of one episode at frame $t+1$.

\begin{enumerate}
  \item Iterate the following steps from $i=1$ to $t$. At each iteration, $(\hat{\alpha}_i,\hat{\nu}_i)$ and $(\alpha_i,\nu_i)$ are the predicted and ground-truth \textit{actions}, respectively, of the HM scanpath for the \textit{viewer}, and $\mathbf{o}_i$ is the \textit{observation} of the FoV content.
  \item Take the \textit{action} of $(\hat{\alpha}_i,\hat{\nu}_i)$ using the DRL network, given the current \textit{observation} $\{\mathbf{o}_1, \ldots, \mathbf{o}_i\}$ till frame $i$. The \textit{action} of $\hat{\alpha}_i$ selects one among 8 discrete HM scanpath directions, i.e., $\{ 0^{\circ}, 45^{\circ}, \cdots, 315^{\circ} \}$. The \textit{action} of $\hat{\nu}_i$ is a scalar of HM scanpath magnitude.
  \item Calculate \textit{rewards} $(r^{\alpha}_{i}, r^{\nu}_{i})$ from the \textit{reward} estimator with \eqref{reward-alpha} and \eqref{reward-nu}, which measures how close the \textit{action} $(\hat{\alpha}_i,\hat{\nu}_i)$ is to the ground-truth HM scanpath $(\alpha_i,\nu_i)$. Here, the sums in \eqref{reward-alpha} and \eqref{reward-nu}  are not required for the \textit{reward} calculation, since the ground-truth HM scanpath of online prediction is from a single \textit{viewer}, rather than from all subjects.
  \item Generate new \textit{observation} ${\mathbf{o}_{i+1}}$ from the FoV extractor with the above \textit{action} $(\hat{\alpha}_i,\hat{\nu}_i)$, and then input it to the DRL network.
  \item Update the DRL network using \eqref{opt-1}, \eqref{opt-2} and \eqref{opt-3} and stop iterations, if the iteration number $i$ is equivalent to $t$. Otherwise, proceed to step 2) for the next iteration.
\end{enumerate}
Here,  the definitions of \textit{action}, \textit{reward} and \textit{observation} are the same as those in Section \ref{train}.
The above iterations share the same implementation of training the DRL model in offline-DHP, which was already presented in Section \ref{train}.

Once the above iterations are terminated, our algorithm moves to the next episode.
After a number of episodes, the training stage ends for frame $t+1$, when meeting the termination conditions.
In our approach, there are two termination conditions. The first condition is the maximum number $E$ of episodes.
The second condition is based on the metric of mean overlap (MO), which measures how close the predicted HM position is to the ground-truth HM position.
MO ranges from 0 to 1, and a larger MO indicates a more precise prediction.
Specifically, MO is defined as,
\begin{equation}
\label{mo-defination}
\textrm{MO} =\frac{A(\textrm{FoV}_{p} \cap \textrm{FoV}_{g})}{A(\textrm{FoV}_{p} \cup \textrm{FoV}_{g})},
\end{equation}
where $\textrm{FoV}_{p}$ and $\textrm{FoV}_{g}$ represent the FoVs at the predicted and ground-truth HM positions, respectively.
In \eqref{mo-defination}, $A$ represents the area of a panoramic region, which accounts for number of pixels.
Then, the MO result of \eqref{mo-defination} at each episode is compared with a threshold $th_{\text{MO}}$ to determine whether the training stage is terminated.

Finally, the trained DRL network can be obtained at frame $t+1$, once satisfying one of  the above termination conditions.
Algorithm \ref{online-DHP-algorithm-training} presents the summary of the training stage in online-DHP.

\begin{algorithm}
   \caption{\hspace{-.3em}: Algorithm for the training stage of online-DHP to predict the HM position at frame $t+1$.}
   \label{online-DHP-algorithm-training}
   \footnotesize
\begin{algorithmic}[1]
   \STATE {\bfseries Input:} Panoramic video frames $\{\mathbf{F}_1, \ldots, \mathbf{F}_t \}$, and the ground-truth HM positions of the \textit{viewer} $\{(x_{1},y_{1}),\ldots, (x_{t},y_{t})\}$.
   \STATE Initialize the DRL network of online-DHP with parameter vectors $\{ \theta_{{\nu}}, \theta_{{\pi}}, \theta_{V} \}$, by loading the network of offline-DHP.
   \FOR{$e=1$ {\bfseries to} $E$}
       \STATE Initialize the HM position to be the center of the front region: $\hat{x}_1=0,\hat{y}_1=0$.
       \STATE Initialize the LSTM feature to be the zero vector: $\mathbf{f}_0=\mathbf{0}$.
       \FOR{$i=1$ {\bfseries to} $t-1$}
           \STATE Extract \textit{observation} $\mathbf{o}_i$ (i.e., FoV) from $\mathbf{F}_i$ according to $(\hat{x}_{i},\hat{y}_{i})$.
           \STATE Obtain \textit{policy} $\pi_{i}$ and LSTM feature $\mathbf{f}_{i}$ using the DRL network with $\{\mathbf{o}_i, \mathbf{f}_{i-1}, \theta_{\pi}\}$.
           \STATE Select \textit{action} $\hat{\alpha}_{i}$ according to the $\epsilon$-greedy policy of $\pi_{i}$.
           \STATE Generate \textit{action} $\hat{\nu}_{i}$ using the DRL network given $\mathbf{o}_{i}, \mathbf{f}_{i-1}$ and $ \theta_{\nu}$.
           \STATE Calculate $(\hat{x}_{i+1}, \hat{y}_{i+1})$ with regard to $\hat{\alpha}_{i},\hat{\nu}_{i}$, and $(\hat{x}_{i}, \hat{y}_{i})$.
           \STATE Estimate \textit{rewards} $r^{\nu}_{i}$ and $ r^{\alpha}_{i}$ through \eqref{reward-alpha} and \eqref{reward-nu} for $(\hat{\alpha}_{i},\hat{\nu}_{i})$ .
           \STATE Calculate the MO between $(\hat{x}_{i},\hat{y}_{i})$ and $(x_{i},y_{i})$, denoted as $\text{MO}_i$.
           \STATE Store a set of experiences: $\{ \mathbf{o}_{i}, \! \mathbf{f}_{i-1},\! \hat{\nu}_{i},\! \hat{\alpha}_{i},\! r^{\nu}_{i},\! r^{\alpha}_{i} \}$.
           \STATE $i \leftarrow i+1$.
       \ENDFOR
       \STATE Update $\{ \theta_{\nu}, \theta_{\pi}, \theta_{V} \}$ according to \eqref{opt-1}, \eqref{opt-2}, \eqref{opt-3}, in which $\{ \theta^{n}_{\nu}, \theta^{n}_{\pi}, \theta^{n}_{V} \}$ are replaced by $\{ \theta_{\nu}, \theta_{\pi}, \theta_{V} \}$.
       \STATE $e \leftarrow e+1$.
       \STATE Calculate the average MO through $\text{MO} =  \frac{\sum_{i=1}^{t-1} \text{MO}_{i}}{t-1}$.
       \IF{$\text{MO}> th_{\text{MO}}$}
           \STATE \textbf{break}
       \ENDIF
  \ENDFOR
  \STATE {\bfseries Return:} The trained parameter vectors: $\{ \theta_{\nu}, \theta_{\pi}, \theta_{V} \}$.
\end{algorithmic}
\end{algorithm}
\vspace{-1em}

\subsection{Stage II: Prediction}
When the average MO is larger than threshold $th_{\text{MO}}$, the switch of Figure \ref{online-framework} is turned to ``predict'', and the DRL network makes an action of the HM scanpath at frame $t+1$.
Note that if the number of training episodes exceeds $E$, then the ``predict'' is also switched on, such that the training episodes end in a limited time.
When entering the prediction stage, the DRL model trained in the first stage is used to produce the HM position as follows.

First, the LSTM features $\{\mathbf{f}_i\}_{i=1}^{t-1}$ are sequentially updated from frame $1$ to $t-1$, based on the observed FoVs $\{\mathbf{o}\}_{i=1}^{t-1}$ and the DRL parameters $\theta_{\pi}$ of the training stage. Note that the LSTM feature is initialized with the zero vector $\mathbf{0}$ at frame $1$. Then, $\{\mathbf{o}_t, \mathbf{f}_{t-1}, \theta_{\pi}\}$ produce action $\hat{\alpha}_t$ of the HM scanpath direction. In addition, the HM scanpath magnitude $\hat{\nu}_t$ is generated using $\{\mathbf{o}_t, \mathbf{f}_{t-1}, \theta_{\nu}\}$, in which the parameters of $\theta_{\nu}$ are obtained at the training stage. Afterwards, the HM position $(\hat{x}_{t+1}, \hat{y}_{t+1})$ at frame $t+1$ can be predicted, given the ground-truth HM position $(\!x_{t}, \!y_{t}\!)$ and the estimated HM scanpath $(\hat{\alpha}_t, \hat{\nu}_t)$ at frame $t$. Algorithm \ref{online-DHP-algorithm-predicting} presents the summary of the prediction stage in online-DHP. Finally, online-DHP is achieved by alternating between the training and prediction stages until the currently processed frame.

\begin{algorithm}
   \caption{\hspace{-.3em}: Algorithm for the prediction stage of online-DHP at frame $t+1$.}
   \label{online-DHP-algorithm-predicting}
   \footnotesize
\begin{algorithmic}[1]
   \STATE {\bfseries Input:} The trained parameter vectors: $\{ \theta_{\nu}, \theta_{\pi}, \theta_{V} \}$ from the training stage, panoramic video frames $\{\mathbf{F}_1, \ldots, \mathbf{F}_t \}$, and the ground-truth HM positions of the \textit{viewer} $\{(x_{1},y_{1}),\ldots, (x_{t},y_{t})\}$.
   \STATE Initialize the LSTM feature with the zero vector: $\mathbf{f}_0=\mathbf{0}$.
   \FOR{$i=1$ {\bfseries to} $t-1$}
       \STATE Extract \textit{observation} $\mathbf{o}_i$ (i.e., FoV) from $\mathbf{F}_i$ according to $(x_{i},y_{i})$.
       \STATE Obtain LSTM feature $\mathbf{f}_{i}$ using the DRL network with $\{\mathbf{o}_{i},\!\mathbf{f}_{i-1},\! \theta_{\pi}\}$.
       \STATE $i \leftarrow i+1$.
   \ENDFOR
   \STATE Extract \textit{observation} $\mathbf{o}_t$ (i.e., FoV) from $\mathbf{F}_t$ according to $(x_{t},y_{t})$.
   \STATE Obtain \textit{policy} $\pi_{t}$ using the DRL network with $\{\mathbf{o}_{t},\!\mathbf{f}_{t-1},\! \theta_{\pi}\}$.
   \STATE Choose \textit{action} $\hat{\alpha}_{t}$ using the greedy policy based on $\pi_{t}$.
   \STATE Generate HM magnitude $\hat{\nu}_{t}$ using the DRL network with $\{\mathbf{o}_{t}, \mathbf{f}_{t-1}, \theta_{\nu}\}$.
   \STATE Estimate HM position $(\!\hat{x}_{t+1}\!,\!\hat{y}_{t+1}\!)$ at frame $\!t+1$, upon $\hat{\alpha}_{t},\hat{\nu}_{t}$ and $(\!x_{t}\!,\!y_{t}\!)$.
   \STATE {\bfseries Return:} The HM position at frame $\!t+1$: $(\hat{x}_{t+1}\!,\!\hat{y}_{t+1})$.
   \end{algorithmic}
\end{algorithm}
\vspace{-1em}

\section{Experimental results}
This section presents the experimental results for validating the effectiveness of our offline-DHP and online-DHP approaches. In Section \ref{sec:settings}, we discuss the settings of both offline-DHP and online-DHP in our experiments. Section \ref{abl_ex} presents the results of ablation experiments. Sections \ref{sec:evaluation_offline} and \ref{sec:evaluation_online} compare the performance of our offline-DHP and online-DHP approaches with those of other approaches in predicting HM positions, in the offline and online scenarios, respectively.

\subsection{Settings}\label{sec:settings}
For evaluating the performance of offline-DHP, we randomly divided all 76 panoramic sequences of our PVS-HM database into a training set (61 sequences) and a test set (15 sequences). In training of the DRL model, the hyperparameters $\rho$, $\varrho$ and $\varsigma$ of \eqref{reward-alpha} and \eqref{reward-nu} were tuned over the training set,  when estimating the \textit{reward} of HM scanpath prediction. As a result, $\rho$, $\varrho$ and $\varsigma$ were set to be $42$, $0.7$ and $1.0$. In addition, we followed \cite{mnih2016asynchronous} to set the other hyperparameters of DRL. For example, we set the discount factor $\gamma$ of \eqref{opt-2} and \eqref{opt-3} to be $0.99$ for \textit{reward} optimization. In our experiments, all 61 training sequences, each of which corresponds to a local DRL network, were used to update the global network as the trained DRL model.
The number of DRL workflows $N$ in the offline-DHP framework was set to be 58, which is the same as the number of subjects in our PVS-HM database.
Similar to \cite{matin1974saccadic},  the HM positions predicted by the 58 DRL workflows were convoluted with a 2D Gaussian filter at each panoramic frame, to generate the HM map.
In our experiments, the HM maps in a panorama were projected to a 2D plane for facilitating visualization.
For evaluation, we measure the prediction accuracy of HM maps in terms of correlation coefficient(CC), normalized scanpath saliency (NSS), and area under receiver operating characteristic curve (AUC), which are three effective evaluation metrics \cite{Li_2015_ICCV} in saliency detection.
Here, the shuffled-AUC is applied, in order to remove the influence of FCB in  the evaluation.
Note that larger values of CC, NSS and shuffled-AUC correspond to a more accurate prediction of HM maps.

For evaluating the performance of online-DHP, we compared our approach with \cite{hu2017deep} and two baseline approaches.
The same as \cite{hu2017deep}, MO of \eqref{mo-defination} is measured as the metric to evaluate the accuracy of online prediction in HM positions. Note that a larger value of MO means a more accurate online prediction in HM positions.
Since the DRL network of offline-DHP was learned over 61 training sequences and used as the initial model of online-DHP, our comparison was conducted on all 15 test sequences of our PVS-HM database. In our experiments, the comparison was further performed over all test sequences of the database presented in \cite{hu2017deep}, in order to test the generalization ability of our online-DHP approach.
In our online-DHP approach, the hyperparameters of $\rho$, $\varrho$, $\varsigma$ and $\gamma$ were set to be the same as those of the DRL workflows of offline-DHP.
The other hyperparameters were identical to those in the most recent DRL work of \cite{mnih2016asynchronous}.
In addition, the maximum number of episodes and the MO threshold were set to be 30 and 0.7, respectively, as the termination conditions in the training stage of online-DHP. Note that the MO threshold ensures the accuracy of HM position prediction, while the maximum episode number constrains the computational time of online-DHP.

\begin{figure}
\vspace{-1em}
	\begin{center}
		\centerline{\includegraphics[width=.6\columnwidth]{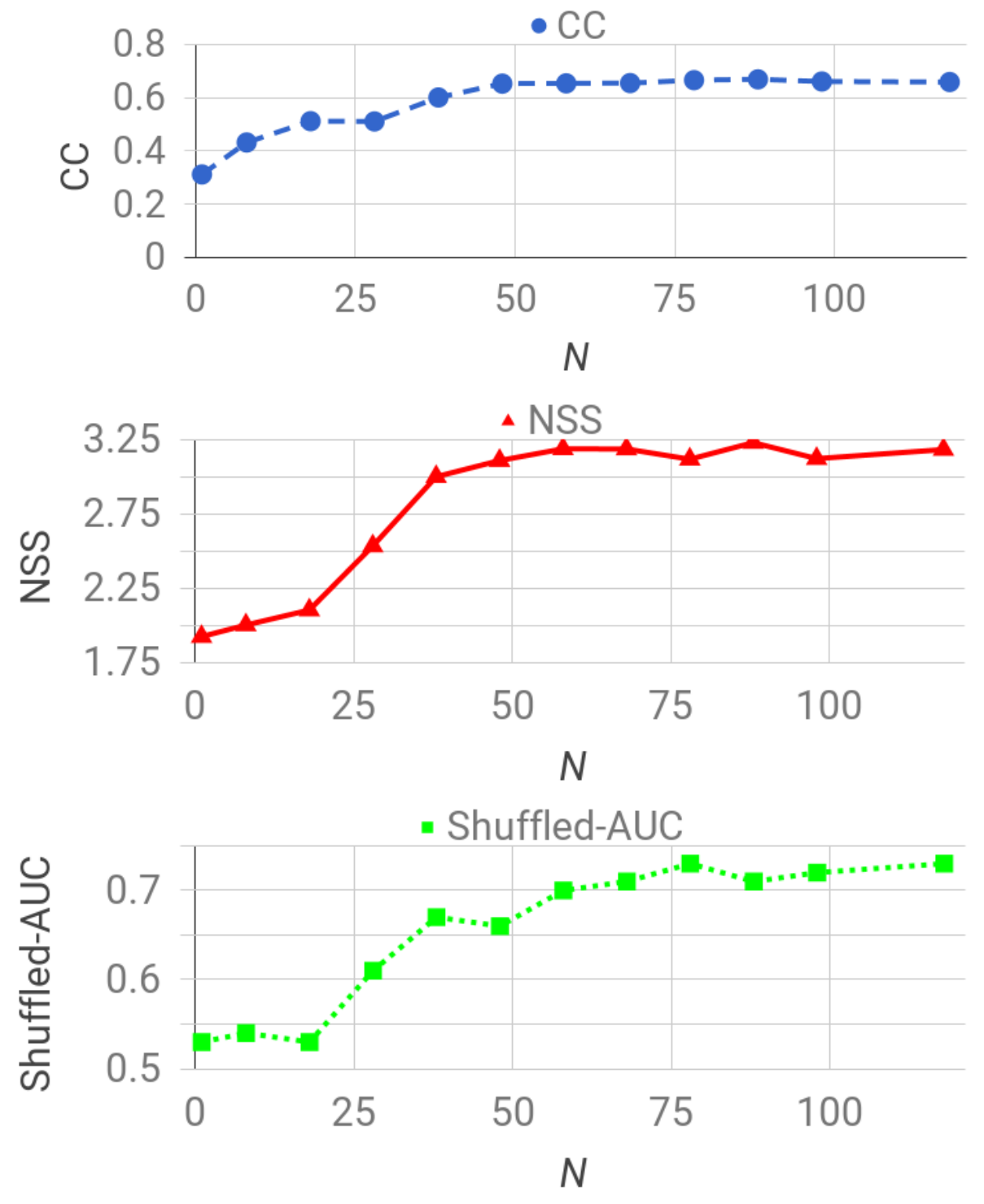}}
        \vspace{-1em}
		\caption{\footnotesize{Performance of offline-DHP at different numbers of workflows.}}
		\label{N_Ablation}
	\end{center}
\vspace{-2em}
\end{figure}

\begin{figure*}
\vspace{-1em}
	\begin{center}
		\centerline{\includegraphics[width=1.8\columnwidth]{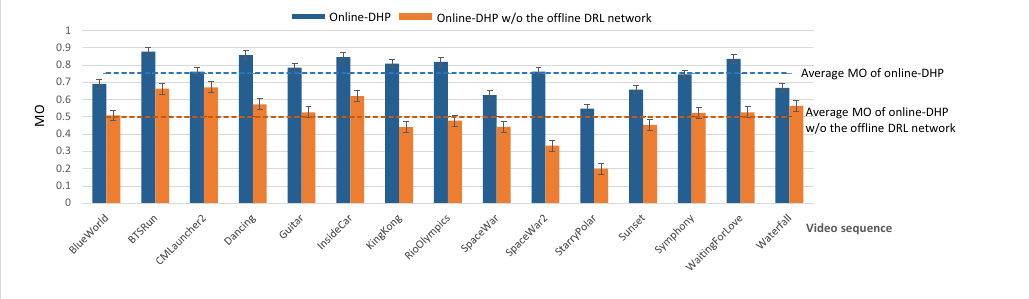}}
      \vspace{-2em}
		\caption{\footnotesize{MO results between the online-DHP approaches with and without the trained offline-DHP network.}}
		\label{Online_compare}
	\end{center}
\vspace{-2em}
\end{figure*}

\begin{figure*}
	\begin{center}
\vspace{-1em}
		\centerline{\includegraphics[width=1.8\columnwidth]{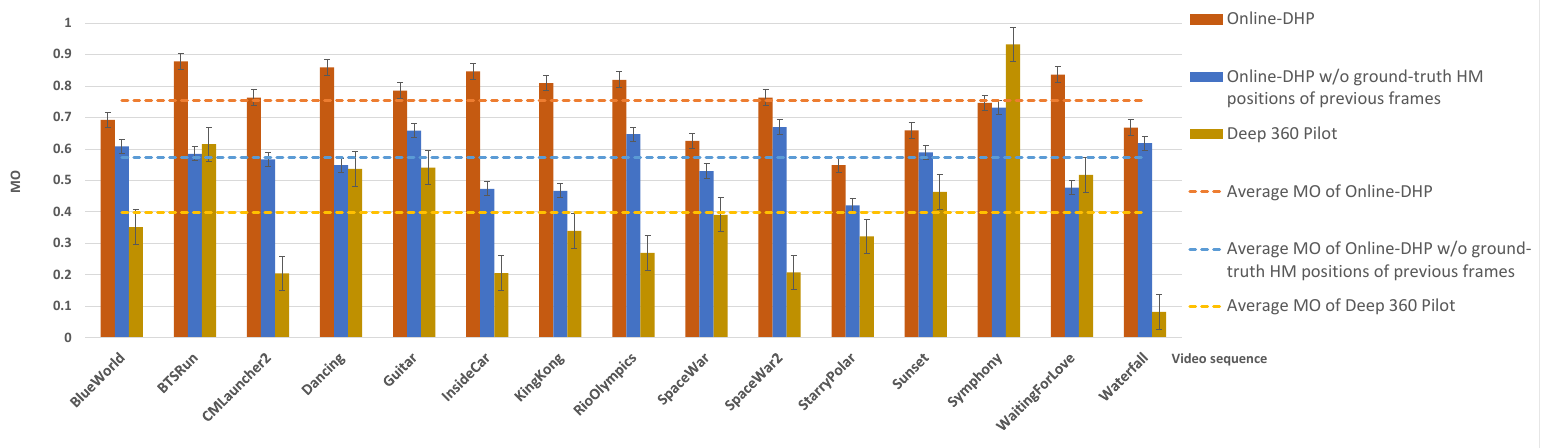}}
		\vspace{-1.5em}
        \caption{\footnotesize{MO results for Deep 360 Pilot,  online-DHP approach, and online-DHP w/o ground- truth HM positions of previous frames.}}
		\label{MO_result_1}
	\end{center}
\vspace{-1em}
\end{figure*}

\begin{table*}
	\vspace{-1em}
	\begin{center}

		\caption{$\Delta$CC, $\Delta$NSS, $\Delta$S-AUC and $\Delta$MO between offline-DHP/online-DHP and the corresponding supervised baseline over 15 test sequences.}
		\label{CC_NSS_MO_table}
		\vspace{-1.5em}

		\tiny

		\resizebox{\textwidth}{!}{

			\begin{tabular}{ccc*{16}c}
				\tabincell{c}{\rotatebox{45}{}} & \rotatebox{45}{}& \rotatebox{45}{}

				& \rotatebox{45}{StarryPolar} & \rotatebox{45}{Symphony} & \rotatebox{45}{SpaceWar} & \rotatebox{45}{RioOlympics} & \rotatebox{45}{InsideCar}

				& \rotatebox{45}{SpaceWar2} & \rotatebox{45}{Sunset} & \rotatebox{45}{BlueWorld} & \rotatebox{45}{Waterfall} & \rotatebox{45}{Dancing}

				& \rotatebox{45}{CMLauncher2} & \rotatebox{45}{Guitar} & \rotatebox{45}{KingKong} & \rotatebox{45}{BTSRun} & \rotatebox{45}{WaitingForLove}

				& \rotatebox{45}{\textbf{Average}}

				\\

				\toprule
                \multirow{3}{*}{Offline}

				&\multicolumn{2}{c}{$\Delta$CC}


				& -0.475 & -0.076 & 0.041 & 0.007 & 0.441 & 0.236 & 0.093 & 0.178 & 0.109 & 0.416 & 0.079 & 0.302 & 0.101 & 0.001 & -0.089 & 0.091

				\\
				&\multicolumn{2}{c}{$\Delta$NSS}


				& -0.524 & 0.834 & 0.534 & 0.566 & 0.500 & 2.625 & 0.735 & 1.469 & 1.014 & 3.768 & 1.013 & 2.342 & 0.242 & 0.813 & 0.062 & 1.066
				\\

				&\multicolumn{2}{c}{$\Delta$S-AUC}


				& -0.025 & 0.024 & -0.025 & 0.156 & 0.330 & 0.112 & -0.056 & 0.113 & 0.025 & 0.267 & 0.255 & 0.040 & 0.133 & 0.320 & 0.101 & 0.118

				\\
                \midrule
                \multirow{1}{*}{Online}&
				\multicolumn{2}{c}{$\Delta$MO}


				& 0.06 & 0.03 & 0.06 & 0.05 & 0.04 & 0.07 & 0.05 & 0.05 & 0.05 & 0.03 & 0.04 & 0.06 & 0.04 & 0.03 & 0.03 & 0.05

				\\
				\bottomrule
			\end{tabular}
		}
	\end{center}
	\vspace{-2em}
\end{table*}

\subsection{Ablation experiments}\label{abl_ex}

\textbf{Ablation on the workflow number in offline-DHP.} Our offline-DHP approach generates the HM maps of panoramic video through the predicted HM positions of multiple workflows. Thus, we conducted the ablation experiments to investigate the performance of offline-DHP at different numbers of workflows.
Figure \ref{N_Ablation} shows the results of CC, NSS and shuffled-AUC for our offline-DHP approach, when the number of workflows $N$ varies from 1 to 118. Note that the results in this figure are averaged over all 15 test panoramic sequences.
We can see from Figure \ref{N_Ablation} that CC approximately converges at $N \geq 48$, and that NSS and shuffled-AUC approximately converge at $N \geq 58$.
Thus, we set the number of workflows $N$ to be 58 in our experiments.

\begin{table*}
\vspace{-.5em}
    \begin{center}

        \caption{CC results of offline HM map prediction by our and other approaches over 15 test sequences.}
        \vspace{-1.5em}
        \label{CC-table}
        \begin{threeparttable}
        \tiny

        \resizebox{\textwidth}{!}{

            \begin{tabular}{cc*{16}{c}c}
                                     \tabincell{c}{\rotatebox{45}{CC}} & \rotatebox{45}{Method}

                                               & \rotatebox{45}{StarryPolar} & \rotatebox{45}{Symphony} & \rotatebox{45}{SpaceWar} & \rotatebox{45}{RioOlympics} & \rotatebox{45}{InsideCar}

                                               & \rotatebox{45}{SpaceWar2} & \rotatebox{45}{Sunset} & \rotatebox{45}{BlueWorld} & \rotatebox{45}{Waterfall} & \rotatebox{45}{Dancing}

                                               & \rotatebox{45}{CMLauncher2} & \rotatebox{45}{Guitar} & \rotatebox{45}{KingKong} & \rotatebox{45}{BTSRun} & \rotatebox{45}{WaitingForLove}

                                               & \rotatebox{45}{\textbf{Average}}

                \\

                \toprule

                \multirow{4}{*}{\rotatebox{45}{Non-FCB}}

                \abovespace

                            & Our

                                     & 0.185 & \textbf{0.710} & \textbf{0.573} & \textbf{0.717} & \textbf{0.783} & \textbf{0.673} & \textbf{0.673} & \textbf{0.678} & \textbf{0.763} & \textbf{0.837} & \textbf{0.585} & \textbf{0.645} & \textbf{0.751} & \textbf{0.764} & \textbf{0.471} & \textbf{0.654}

                            \\

                            & BMS

                                     & \textbf{0.450} & 0.167 & 0.274 & 0.228 & 0.331 & 0.067 & 0.463 & 0.169 & 0.393 & 0.121 & 0.203 & 0.328 & 0.105 & 0.105 & 0.223 & 0.242

                            \\

                            & OBDL

                                     & 0.107 & 0.184 & 0.028 & 0.190 & 0.260 & 0.100 & 0.308 & 0.027 & 0.025 & 0.176 & 0.117 & 0.066 & 0.125 & 0.047 & 0.222 & 0.132

                            \\

                            \belowspace

                            & SALICON $^{\ast}$

                                     & 0.168 & 0.216 & 0.106 & 0.189 & 0.292 & 0.291 & 0.235 & 0.255 & 0.393 & 0.281 & 0.220 & 0.365 & 0.217 & 0.285 & 0.288 & 0.253

                \\

                \midrule

                \multirow{4}{*}{\rotatebox{45}{FCB}}

                \abovespace

                            & Our

                                     & 0.497 & \textbf{0.816} &  \textbf{0.574} &  \textbf{0.768} &  \textbf{0.712} &  \textbf{0.655} & \textbf{0.810} &  \textbf{0.748} & \textbf{0.797} &  \textbf{0.764} &  \textbf{0.747} & \textbf{0.652} &  \textbf{0.673} &  \textbf{0.679} &  \textbf{0.677} & \textbf{0.704}

                            \\

                            & BMS

                                     &  \textbf{0.692} & 0.567 & 0.520 & 0.494 & 0.495 & 0.368 & 0.711 & 0.500 & 0.655 & 0.414 & 0.546 & 0.494 & 0.311 & 0.322 & 0.503 & 0.506

                            \\

                            & OBDL

                                         & 0.510 & 0.540 & 0.321 & 0.441 & 0.496 & 0.455 & 0.638 & 0.464 & 0.434 & 0.408 & 0.468 & 0.461 & 0.410 & 0.288 & 0.598 & 0.462

                                \\

                \belowspace

                            & SALICON

                              & 0.642 & 0.670 & 0.552 & 0.629 & 0.539 & 0.527 & 0.745 & 0.530 & 0.621 & 0.453 & 0.651 & 0.496 & 0.445 & 0.431 & 0.622 & 0.570

                            \\

                \midrule

                            \multicolumn{2}{c}{FCB Only}

                                     \abovespace\belowspace

                                     & 0.557 & 0.747 & 0.317 & 0.403 & 0.292 & 0.239 & 0.585 & 0.477 & 0.583 & 0.387 & 0.735 & 0.356 & 0.271 & 0.201 & 0.497 & 0.443

                            \\

                \bottomrule

            \end{tabular}

        }

        \begin{tablenotes}
            \item[] $\ast$ DNN based method has been fine-tuned by our database with their default settings.
        \end{tablenotes}
     \end{threeparttable}

    \end{center}

\end{table*}

\begin{table*}

    \begin{center}
\vspace{-1.5em}
        \caption{NSS results of offline HM map prediction by our and other approaches  over 15 test sequences.}
\vspace{-1em}
        \label{NSS-table}

        \tiny

        \resizebox{\textwidth}{!}{

            \begin{tabular}{cc*{16}{c}c}

                                     \tabincell{c}{\rotatebox{45}{NSS}} & \rotatebox{45}{Method}

                                               & \rotatebox{45}{StarryPolar} & \rotatebox{45}{RioOlympics} & \rotatebox{45}{SpaceWar2} & \rotatebox{45}{Symphony} & \rotatebox{45}{SpaceWar}

                                               & \rotatebox{45}{Waterfall} & \rotatebox{45}{Sunset} & \rotatebox{45}{BlueWorld} & \rotatebox{45}{Guitar} & \rotatebox{45}{Dancing}

                                               & \rotatebox{45}{InsideCar} & \rotatebox{45}{CMLauncher2} & \rotatebox{45}{WaitingForLove} & \rotatebox{45}{BTSRun} & \rotatebox{45}{KingKong}

                                               & \rotatebox{45}{\textbf{Average}}

                \\

                \toprule

                \multirow{4}{*}{\rotatebox{45}{Non-FCB}}

                \abovespace

                            & Our

                                     & 0.899 & \textbf{2.806} & \textbf{2.237} & \textbf{3.346} & \textbf{2.180} & \textbf{3.765} & \textbf{2.529} & \textbf{3.196} & \textbf{3.461} & \textbf{5.297} & \textbf{4.402} & \textbf{3.529} & \textbf{2.278} & \textbf{4.572} & \textbf{3.334} & \textbf{3.189}

                            \\

                            & BMS

                                     & \textbf{1.313} & 0.772 & 0.137 & 0.710 & 0.807 & 1.673 & 1.613 & 0.841 & 1.497 & 0.670 & 1.657 & 1.034 & 0.997 & 0.546 & 0.119 & 0.959

                            \\

                            & OBDL

                                     & 0.126 & 0.637 & 0.301 & 0.260 & 0.064 & 0.073 & 1.015 & 0.035 & 0.393 & 0.980 & 1.375 & 0.660 & 0.964 & 0.215 & 0.107 & 0.480

                            \\

                            \belowspace

                            & SALICON

                                      & 0.628 & 0.584 & 0.396 & 1.093 & 1.348 & 1.528 & 1.194 & 0.877 & 1.167 & 1.541 & 0.876 & 1.265 & 0.858 & 1.121 & 1.362 & 1.056

                            \\

                \midrule

                \multirow{4}{*}{\rotatebox{45}{FCB}}

                \abovespace

                            & Our

                                     & 1.825 & \textbf{2.911} & \textbf{2.064} & \textbf{3.756} & \textbf{2.031} & \textbf{3.755} & \textbf{2.943} & \textbf{3.393} & \textbf{3.395} & \textbf{4.608} & \textbf{3.816} & \textbf{4.463} & \textbf{3.351} & \textbf{3.931} & \textbf{2.883} & \textbf{3.275}

                            \\

                            & BMS

                                     & \textbf{2.206} & 1.779 & 1.063 & 2.537 & 1.667 & 2.891 & 2.507 & 2.280 & 2.386 & 2.366 & 2.508 & 3.136 & 2.434 & 1.771 & 1.288 & 2.188

                            \\

                            & OBDL

                                     & 1.712 & 1.572 & 1.371 & 2.368 & 1.055 & 1.920 & 2.225 & 2.007 & 2.377 & 2.319 & 2.556 & 2.777 & 2.912 & 1.580 & 1.693 & 2.030

                            \\

                            \belowspace

                            & SALICON

                            & 2.008 & 2.219 & 1.503 & 2.799 & 1.669 & 2.736 & 2.522 & 2.218 & 2.385 & 2.568 & 2.794 & 3.766 & 3.038 & 2.358 & 1.709 & 2.419

                            \\

                \midrule

                            \multicolumn{2}{c}{FCB Only}

                                     \abovespace\belowspace

                                             & 2.388 & 1.613 & 0.699 & 4.123 & 1.190 & 3.191 & 2.406 & 2.286 & 1.828 & 2.151 & 1.387 & 5.764 & 2.600 & 1.095 & 1.020 & 2.249

                            \\

                \bottomrule

            \end{tabular}

        }

    \end{center}
\vspace{-1em}
\end{table*}

\textbf{Reinforcement learning vs. supervised learning.}
Here, we evaluate the effectiveness of reinforcement learning applied in our approach by comparing with the supervised learning baseline.
In the supervised learning baseline, the reinforcement learning component of our approach is replaced by a regressor and classifier.
Specifically, the input to the supervised learning baseline is the FoV at the current frame, the same as our DHP approach. Then, the supervised learning baseline predicts the continuous magnitude of HM scanpath through a regressor.
Additionally,  the supervised learning baseline incorporates a classifier to predict the HM scanpath direction among 8 discrete directions in GCS: $\{ 0^{\circ}, 45^{\circ}, \cdots, 315^{\circ} \}$.
This ensures that the output of the baseline is the same as that of our DHP approach.
For fair comparison, the DNN architecture of our DHP approach is used as the regressor and classifier, which have the same convolutional layers and LSTM cells as our approach.
The magnitude regressor is trained by the MSE loss function,
while the direction classifier is trained by the cross entropy loss function.

First, we compare the supervised learning baseline with our offline-DHP approach.
To this end, the same as our offline-DHP approach, the supervised learning baseline runs 58 workflows to predict different HM positions for each panoramic frame.
In each workflow, the baseline randomly samples one direction for the possible HM scanpath at each frame, according to the probabilities of directions by the trained classifier.
Then, several HM positions are obtained upon the HM directions of all workflows, given the magnitude predicted by the trained regressor.
Finally, the HM map is produced by convoluting these HM positions.
Table \ref{CC_NSS_MO_table} reports the CC, NSS and shuffled-AUC increase ($\Delta$CC, $\Delta$NSS and $\Delta$S-AUC) of our offline-DHP approach with the supervised learning approach as an anchor.
We can see that the proposed offline-DHP approach performs much better against the supervised learning baseline.
This validates the effectiveness of reinforcement learning applied in offline-DHP.

Second, we compare the supervised learning baseline with our online-DHP approach.
The baseline predicts the HM position at the next frame using the trained magnitude regressor and direction classifier.
In contrast, our online-DHP approach predicts HM positions, based on reinforcement learning as introduced in Section 5.
Table \ref{CC_NSS_MO_table} tabulates the MO improvement ($\Delta$MO) of our online-DHP approach over the supervised learning baseline.
As seen in this table, our online-DHP approach outperforms the supervised learning baseline in all sequences.
Therefore, reinforcement learning is also effective in online-DHP.

\textbf{Influence of offline DRL network to online-DHP.} It is interesting to analyze the benefits of incorporating the DRL network of offline-DHP in our online-DHP approach, since the online-DHP approach is based on the offline DRL network. Figure \ref{Online_compare} shows the MO results of our online-DHP approach with and without the offline DRL network. As observed in this figure, the offline DRL network is able to increase the MO results of our online-DHP approach, for all 15 sequences. In addition, the MO value can be increased from 0.50 to 0.75 on average, when the offline DRL network is incorporated in online-DHP. Therefore, the learned DRL network of offline-DHL also benefits the online prediction of HM positions in online-DHL.

\textbf{Performance of online-DHP w/o previous ground-truth HM positions.} For each test sequence, our online-DHP takes as input the ground-truth HM positions of previous frames to predict subsequent HM positions.
The online-DHP approach belongs to online machine learning, and it is opposed to batch learning of deep 360 pilot \cite{hu2017deep}, which generates the predictor by learning on the entire training dataset at once.
Note that there is no online machine learning approach for predicting HM positions, and we can only compare with deep 360 pilot.
For fair comparison with deep 360 pilot, Figure \ref{MO_result_1} shows the results of our online-DHP approach using previous predicted HM positions as input, i.e., online-DHP w/o ground-truth HM positions of previous frames.
As observed in Figure \ref{MO_result_1}, our online-DHP approach (MO = 0.57) performs considerably better than deep 360 pilot (MO = 0.40), when the previous ground-truth HM positions are not available in these two approaches for fair comparison.
In addition, the ground-truth HM positions of previous frames can improve the performance of online-DHP, with MO increasing from 0.57 to 0.75 on average.

\subsection{Performance evaluation on offline-DHP}\label{sec:evaluation_offline}
\label{compare}

\begin{figure*}
\vspace{-1em}
	\begin{center}
		\centerline{\includegraphics[width=2\columnwidth]{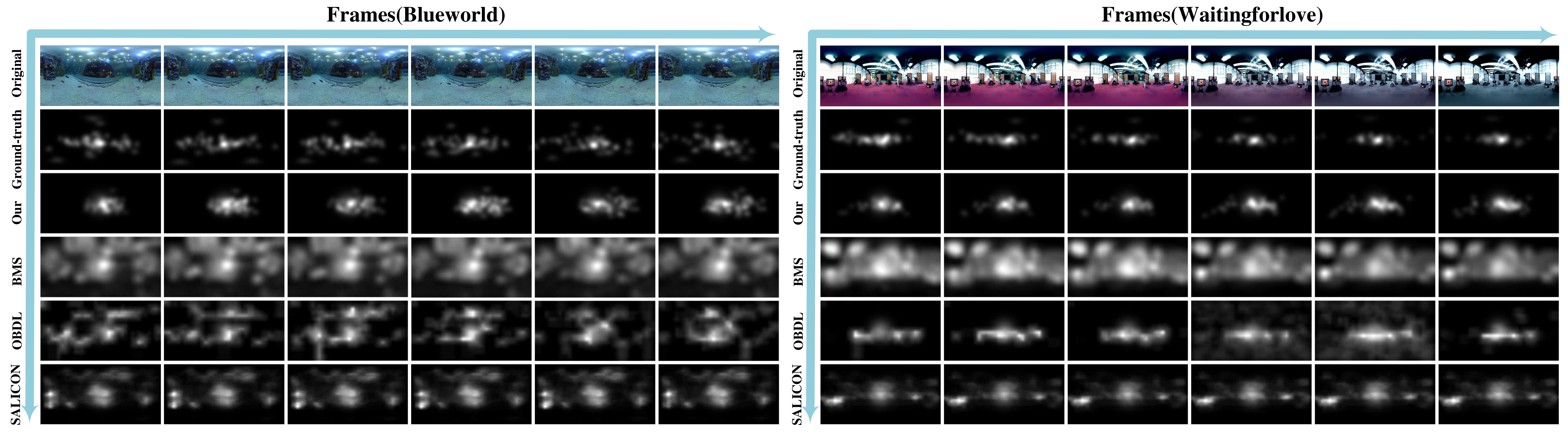}}
\vspace{-1em}
                  \caption{\footnotesize{HM maps of several frames selected from two test sequences in our PVS-HM database. They are all visualized in the 2D coordination.  The second row shows the ground-truth HM maps, which are generated upon the HM positions of all 58 subjects. The third to sixth rows show the HM maps of our, BMS \cite{zhang2016exploiting} , OBDL \cite{hossein2015many}, and SALICON \cite{huang2015salicon} approaches. }}
		\label{figure-object}
	\end{center}
\vspace{-2em}
\end{figure*}

\begin{table*}

    \begin{center} 
        \caption{Shuffled-AUC results of HM map prediction by our and other approaches (without FCB) over 15 test sequences.}\label{Table-s-AUC}
\vspace{-1.2em}
        \label{table-result}

        \tiny

        \resizebox{\textwidth}{!}{

            \begin{tabular}{cc*{15}{c}c}

                                             & \rotatebox{45}{Method}

                                             & \rotatebox {45} {KingKong} & \rotatebox {45} {SpaceWar2} & \rotatebox {45} {StarryPolar} & \rotatebox {45} {Dancing} & \rotatebox {45} {Guitar}

                                             & \rotatebox {45} {BTSRun} & \rotatebox {45} {InsideCar} & \rotatebox {45} {RioOlympics} & \rotatebox {45} {SpaceWar} & \rotatebox {45} {CMLauncher2}

                                             & \rotatebox {45} {Waterfall} & \rotatebox {45} {Sunset} & \rotatebox {45} {BlueWorld} & \rotatebox {45} {Symphony} & \rotatebox {45} {WaitingForLove}

                                             & \rotatebox{45}{\textbf{Average}}

                \\

                \toprule


                            & Our

                                      & \textbf{0.72}  & 0.63  & 0.46  & \textbf{0.82}  & \textbf{0.73}  & 0.84  & 0.80  & \textbf{0.69}  & 0.60  & 0.76  & \textbf{0.72}  & 0.64  & \textbf{0.70}  & 0.70  & 0.68  & \textbf{0.70}

                            \\

                            & SALICON

                                     & 0.62 & 0.57 & 0.42 & 0.66 & 0.62 & 0.76 & 0.56 & 0.54 & \textbf{0.64} & \textbf{0.79} & 0.66 & 0.62 & 0.65 & 0.70 & 0.71 & 0.64

                            \\

                            & BMS

                                      & 0.65  & \textbf{0.65}  & 0.43  & 0.74  & 0.54  & \textbf{0.88}  & \textbf{0.83}  & 0.53  & 0.62  & 0.63  & 0.66  & \textbf{0.70}  & 0.49  & \textbf{0.77}  & 0.69  & 0.65

                            \\

                            & OBDL

                                     & 0.70  & 0.60  & \textbf{0.55}  & 0.81  & 0.68  & 0.86  & 0.62  & 0.68  & 0.57  & 0.56  & 0.47  & 0.69  & 0.47  & 0.75  & \textbf{0.74}  & 0.65

                            \\

                \bottomrule

            \end{tabular}
        }

    \end{center}
\vspace{-1.5em}
\end{table*} 

\begin{table*}
    \begin{center}
        \caption{MO results of online HM position prediction by our and other approaches.}
        \vspace{-1.5em}
        \label{table-result}
        \begin{threeparttable}
        \tiny
        \resizebox{\textwidth}{!}{
            \begin{tabular}{cc*{16}{c}c}

                            \tabincell{c}{\rotatebox{45}{Method}}
                                   & \rotatebox{45}{KingKong} & \rotatebox{45}{SpaceWar2} & \rotatebox{45}{StarryPolar} & \rotatebox{45}{Dancing} & \rotatebox{45}{Guitar}
                                   & \rotatebox{45}{BTSRun} & \rotatebox{45}{InsideCar} & \rotatebox{45}{RioOlympics} & \rotatebox{45}{SpaceWar} & \rotatebox{45}{CMLauncher2}
                                   & \rotatebox{45}{Waterfall} & \rotatebox{45}{Sunset} & \rotatebox{45}{BlueWorld} & \rotatebox{45}{Symphony} & \rotatebox{45}{WaitingForLove} & \rotatebox{45}{{\textbf{Average}}}

                \\
               \toprule
               \multirow{1}{*}{\rotatebox{0}{Online$^{\ast}$}}
               \abovespace

                           & \bf{0.81} & \bf{0.76} & \bf{0.55} & \bf{0.86} & \bf{0.79} & \bf{0.88} & \bf{0.85} & \bf{0.82} & \bf{0.63} & \bf{0.76} & \bf{0.67} & \bf{0.66} & \bf{0.69} & 0.75 & \bf{0.84} & \bf{0.75}
                         \\
                \multirow{1}{*}{\rotatebox{0}{Deep 360 Pilot}}
                           & 0.34 & 0.21 & 0.32 & 0.54 & 0.54 & 0.62 & 0.21 & 0.27 & 0.39 & 0.21 & 0.08 & 0.46 & 0.35 & \bf{0.93} & 0.52 & 0.40
                         \\
                 \multirow{1}{*}{\rotatebox{0}{Baseline 1$^{\ast}$}}
                          & 0.20 & 0.21 & 0.16 & 0.22 & 0.20 & 0.21 & 0.22 & 0.20 & 0.21 & 0.21 & 0.20 & 0.20 & 0.21 & 0.20 & 0.21 & 0.20
                         \\
                         \belowspace
                 \multirow{1}{*}{\rotatebox{0}{Baseline 2$^{\ast}$}}
                          & 0.22 & 0.23 & 0.20 & 0.22 & 0.23 & 0.24 & 0.23 & 0.23 & 0.22 & 0.25 & 0.25 & 0.21 & 0.23 & 0.22 & 0.23 & 0.23
                         \\
                \bottomrule
            \end{tabular}%
        }
        \begin{tablenotes}
            \item[] $\ast$ Both the online-DHP approach and baseline make prediction based on the ground-truth of previous frames.
         \vspace{-1em}
        \end{tablenotes}
        \end{threeparttable}
        \end{center}
                \vspace{-1em}
\end{table*}

Now, we evaluate the performance of our offline-DHP approach in predicting the HM maps of all 15 test sequences from the PVS-HM database. To the best of our knowledge, there is no work on predicting the HM maps of panoramic video, and saliency prediction is the closest field.
Therefore, we compare our offline-DHP approach to three state-of-the-art saliency detection approaches:  OBDL  \cite{hossein2015many}, BMS \cite{zhang2016exploiting} and SALICON \cite{huang2015salicon}, which are applied to panoramic frames mapped from sphere to plane using equirectangular projection. In particular, OBDL and BMS are the latest saliency detection approaches for videos and
images, respectively. SALICON is a state-of-the-art DNN approach for saliency detection. For fair comparison, we retrained the DNN model of SALICON by fine-tuning over the training set of our database. Note that OBDL and BMS were not retrained because they are not trainable.
In addition to the above three approaches, we also compare our approach to the FCB baseline, since \textit{Finding 1} argues that human attention normally biases toward the front-center regions of panoramic video.
Here, we model FCB using a 2D Gaussian distribution, similar to the center bias of saliency detection.
Appendix A presents the details of the FCB modeling.
In the field of saliency detection, the center bias  \cite{borji2013state} is normally combined with saliency maps to improve the saliency detection accuracy. Hence, we further report the results of HM maps combined with the FCB feature, for our and other approaches.
See Appendix A for more details about the combination of FCB.

Tables \ref{CC-table} and \ref{NSS-table} tabulate the results of CC and NSS in predicting the HM maps of 15 test sequences, for our and other approaches.
In these tables, the results of CC and NSS are averaged over all frames for each test sequence.
As shown in this table, when FCB is not integrated, our offline-DHP approach performs best among all three approaches and the FCB baseline, in terms of CC and NSS.
More importantly, once integrated with FCB, all three approaches have performance improvement, and our approach still performs considerably better than other approaches.
Specifically, our offline-DHP approach increases the average CC value by 0.242, 0.198 and 0.134, compared with OBDL, BMS and SALICON, respectively.
Additionally, the increase of average NSS value is 1.245, 1.087 and 0.856 in our approach, in comparison with OBDL, BMS and SALICON.
In a word, our offline-DHP approach is effective in predicting the HM maps of panoramic video, much better than other approaches and the FCB baseline.

Additionally, Table \ref{Table-s-AUC} compares the performance of our and other approaches in terms of shuffled-AUC.
Note that FCB is not embedded in all approaches, since the shuffled-AUC metric is immune to FCB.
In terms of the average shuffled-AUC, our approach has better performance than other approaches.
This indicates that even not considering the influence of FCB, our approach again outperforms other approaches.
It is worth mentioning that the shuffled-AUC of our offline-DHP approach ranks top in 6 out of 15 test sequences, while SALICON, BMS and ODBL have highest shuffled-AUC in 2, 5 and 2 sequences, respectively.
The probable reasons are as follows. (1) In the evaluation, shuffled-AUC removes the influence of FCB, which can be learned by our offline-DHP approach. (2)  The shuffled-AUC can be high even when the HM maps are non-sparse, i.e., far from ground truth. However, our approach yields more sparse HM maps than other approaches, close to ground-truth (see Figure \ref{figure-object}).

Next, we compare the subjective results. Figure \ref{figure-object} shows several frames from two selected sequences and their ground-truth HM maps.
In Figure \ref{figure-object}, we further visualize the HM maps generated by our and other approaches. Here, the predicted HM maps are integrated with FCB, since the FCB feature can improve the performance of all three approaches (as presented in Table \ref{table-result}).
From this figure, one can observe that the HM maps of our approach are considerably closer to the ground-truth HM maps, compared with other approaches.
This result indicates that our offline-DHP approach is capable of better locating the HM positions of different subjects on panoramic video.

\begin{figure*}
	\begin{center}
		\centerline{\subfigure[Dancing]{\includegraphics[width=1.5\columnwidth]{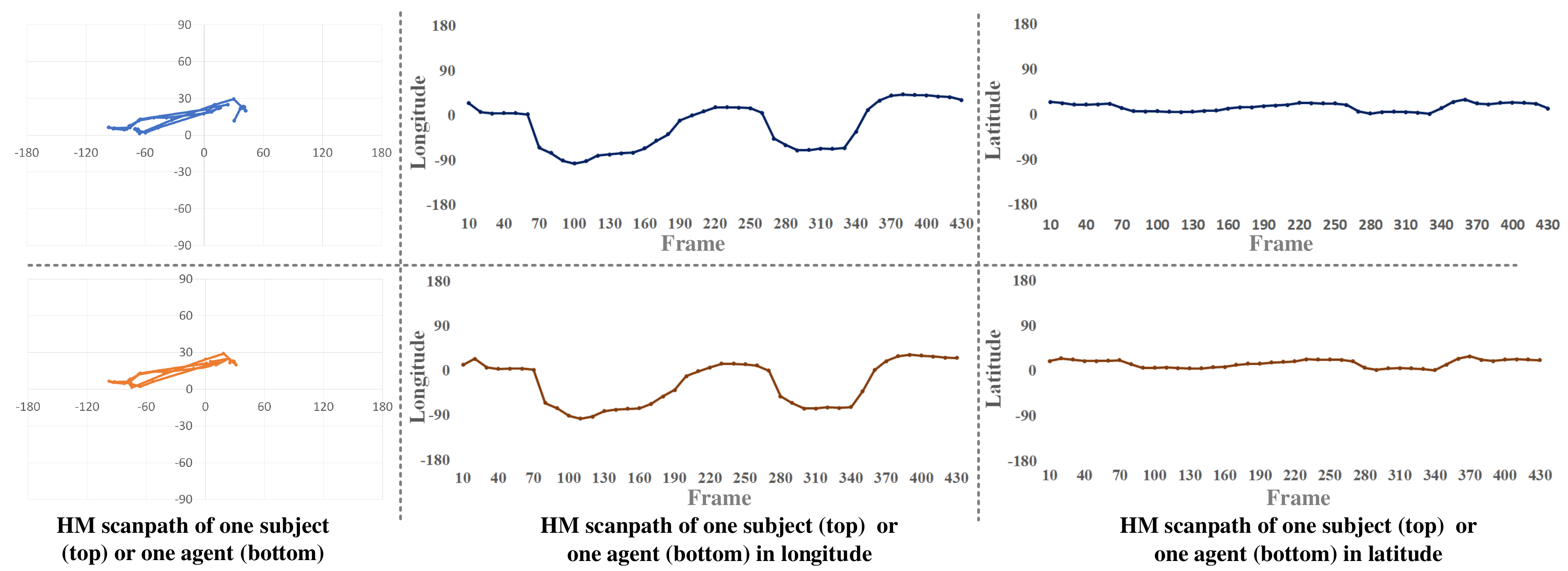}}}
\vspace{-1em}
        \centerline{\subfigure[KingKong]{\includegraphics[width=1.5\columnwidth]{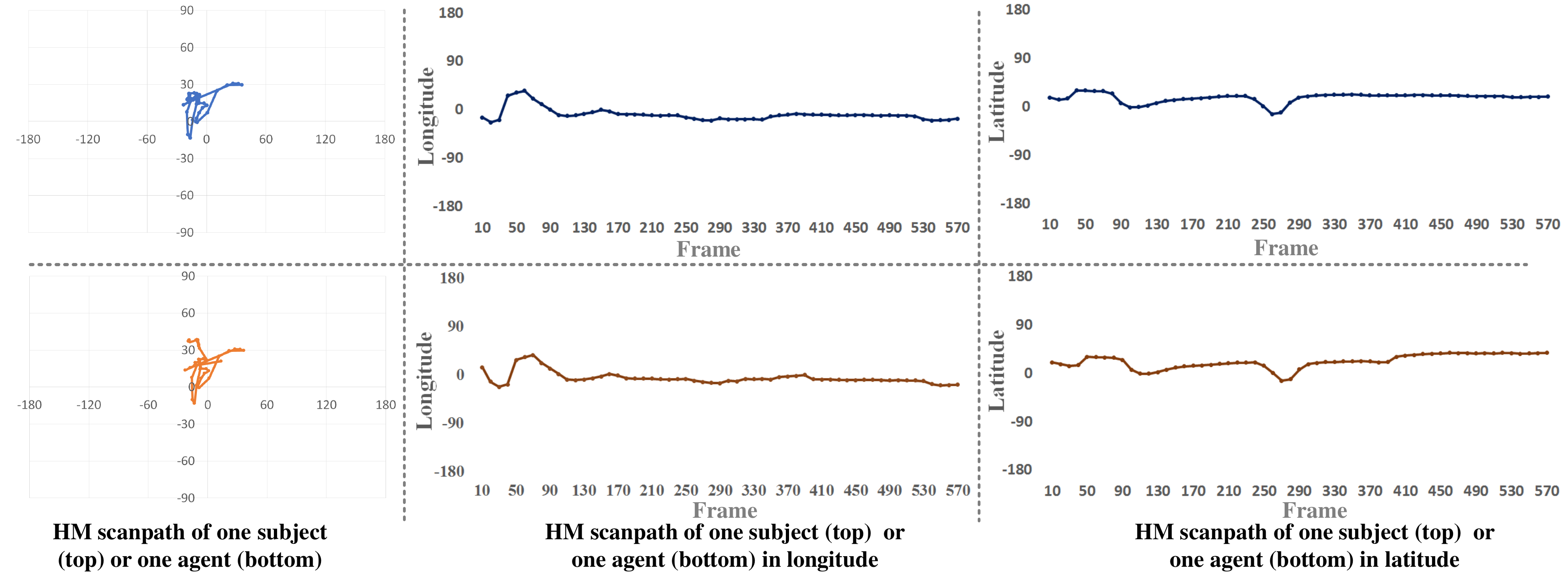}}}
\vspace{-1em}
                   \caption{\footnotesize{Visualization in HM scanpaths generated by one subject and the online-DHP approach, for sequences \textit{Dancing} and \textit{KingKong}. Note that the HM scanpaths of one subject (among 58 subjects) are randomly selected and plotted, and then the corresponding HM scanpaths predicted by online-DHP are plotted.}}
		\label{scan-path-example}
	\end{center}
\end{figure*}

\subsection{Performance evaluation on online-DHP}\label{sec:evaluation_online}
\label{online-compare}

This section evaluates the performance of our online-DHP approach for predicting HM positions in the online scenario.
The online scenario refers to predicting the HM position of one subject at each panoramic frame based on the observed HM positions of this subject at the previous frames.
In our experiments, we compare the performance of online-DHP with the state-of-the-art deep 360 pilot \cite{hu2017deep}, which is the only existing approach for the online prediction of HM positions in panoramic video.
We also compare our online-DHP approach with two baselines. The first baseline (called baseline 1) keeps the HM scanpath of the current frame the same as that at the previous frame, such that the online HM position at each frame can be generated.
The second baseline (called baseline 2) produces the HM positions, using the randomly generated HM scanpaths.

Table \ref{table-result} compares the MO results of our and other approaches for the 15 test sequences of our PVS-HM database. Note that the MO results of each sequence are averaged over the predicted HM positions of all 58 subjects in our database. As observed in this table, our online-DHP approach is significantly superior to two baselines, indicating the effectiveness of applying DRL to predict HM positions online.
Table \ref{table-result} also shows that our online-DHP approach performs considerably better than the deep 360 pilot \cite{hu2017deep}, with an increase of 0.35 in average MO.
In addition, as shown in Table \ref{table-result}, our approach outperforms \cite{hu2017deep} over almost all sequences.
The performance improvement of our approach is because (1) the online DRL model of our approach is capable of generating the accurate \textit{actions} of HM scanpaths, and (2) the DRL network of offline-DHP is incorporated in our online prediction as the prior knowledge. Moreover, our approach is also effective for the generic panoramic sequences, while \cite{hu2017deep} fails in scenery panoramic video. For example, the MO result of \cite{hu2017deep} for the sequence Waterfall is $0.08$, which is far less than $0.67$ MO of online-DHP. This result is primarily because the deep 360 pilot \cite{hu2017deep} relies heavily on the object detection of RCNN.

Moreover, we visualize the ground-truth and predicted HM scanpaths, for subjective evaluation.
Specifically, Figure \ref{scan-path-example} plots the HM scanpaths by one subject and by the online-DHP approach, for the panoramic sequences of Dancing and KingKong.
As shown in this figure, online-DHP is able to obtain similar scanpaths as the subject, such that the HM positions can be accurately predicted online for each panoramic frame.
In conclusion, our subjective evaluation, together with the above objective MO comparison, illustrates that the proposed online-DHP approach is effective in predicting the HM positions with the online manner.

To test the generalizability of our approach, we further evaluate the performance of our, the deep 360 pilot \cite{hu2017deep} and the two baseline approaches on the  sports-360 dataset of \cite{hu2017deep}. For this evaluation, our online-DHP is still based on our offline DRL network that is learned from the training sequences of our PVS-HM database. The MO results are presented in Table \ref{table-result-on360}. From this table, one may observe that our online-DHP approach again outperforms \cite{hu2017deep} and the two baselines, despite testing on the test set of \cite{hu2017deep}. In particular, the MO result of our approach is 0.90  for the panoramic sequences of dance, with 0.11 MO increase over \cite{hu2017deep}. Additionally, our approach improves the MO results of \cite{hu2017deep} by 0.07, 0.01, 0.01, and 0.10, for the sequences of skateboarding, parkour, BMX and basketball, respectively. In other words, the online-DHP approach is superior to the state-of-the-art approach \cite{hu2017deep} in the online prediction of HM positions, over almost all classes of panoramic video sequences.  Therefore, the generalization capability of our online-DHP approach can be confirmed.

\begin{table}
	\begin{center}
		\caption{MO results for online prediction of HM positions over the sports-360 dataset.}
		\vspace{-1em}
		\label{table-result-on360}
		\tiny
		\resizebox{.47\textwidth}{!}{
			\begin{tabular}{c c c c c c}
				
				Method
				& Skateboarding & Parkour & BMX & Dance & Basketball
			
				\\
				\toprule
				DHP
				\abovespace
				
				& \bf{0.78} & \bf{0.75} & \bf{0.72} & \bf{0.90} & \bf{0.77}
				\\
				Deep 360 Pilot
				& 0.71 & 0.74 & 0.71 & 0.79 & 0.67
				\\
				Baseline 1
				& 0.15 & 0.17 & 0.16 & 0.17 & 0.17
				\\
				
				Baseline 2
				& 0.22 & 0.19 & 0.18 & 0.22 & 0.18
				\\
				\bottomrule
			\end{tabular}
		}
	\end{center}
	\vspace{-2em}
\end{table}

\section{Conclusion}
In this paper, we have proposed the DHP approach for predicting HM positions on panoramic video. First, we established a new database named PVS-HM, which includes the HM data of 58 subjects viewing 76 panoramic sequences. We found from our database that the HM positions are highly consistent across humans. Thus, the consistent HM positions on each panoramic frame can be represented in the form of an HM map, which encodes the possibility of each pixel being the HM position. Second, we proposed the offline-DHP approach to estimate HM maps in an offline manner. Specifically, our offline-DHP approach leverages DRL to make decisions on \textit {actions} of HM scanpaths, via optimizing the \textit{reward} of imitating the way that humans view panoramic video.
Subsequently, the HM scanpaths of several \textit{agents} from multiple DRL workflows are integrated to obtain the final HM maps.
Third, we developed the online-DHP approach, which predicts the HM positions of one subject online. In online-DHP, the DRL algorithm was developed to determine the HM positions of one \textit{agent} at the incoming frames, given the \textit{observation} of previous HM scanpaths and the current video content. The DRL algorithm is based on the learned model of offline-DHP in extracting the spatio-temporal features of attention-related content. Finally, the experimental results showed that both offline-DHP and online-DHP are superior to other conventional approaches, in the offline and online tasks of HM prediction for panoramic video.

Humans always perceive the world around them in a panorama, rather than a 2D plane. Therefore, modeling attention on panoramic video is an important component in establishing human-like computer vision systems. It is an interesting future work to apply imitation learning for modeling attention on panoramic video. In particular, the \textit{reward} of DHP may be learned from ground-truth HM data, belonging to inverse reinforcement learning that is a main category of imitation learning.
Moreover, our work at the current stage mainly focuses on predicting HM positions, as the first step toward attention modeling of panoramic video. Future work should further predict eye fixations within the FoV regions of panoramic video. The potential applications of our approach are another promising work in future. For example, the online-DHP approach may be embedded in robotics, to mimic the way in which humans perceive the real world. Besides, panoramic video has large perceptual redundancy, since most of the panoramic regions cannot be seen by humans. It is thus possible to use the offline-DHP approach to remove such perceptual redundancy, and then the bit-rates of panoramic video coding can be dramatically saved.

\appendices
\section{Analysis of FCB Combined in HM Maps}

The saliency detection literature \cite{judd2009learning} has argued that human attention has strong center bias in images or videos, and that the incorporation of center bias can improve the performance of saliency detection. Similarly, FCB exists when viewing panoramic video, as discussed in \textit{Finding 1}. Hence, this appendix presents the combination of the FCB feature and the offline-DHP approach. Here, we apply the FCB feature as an additional channel in generating the HM maps of panoramic video. Specifically, assume that $\mathbf{H}^f$ is the HM map generated by the channel of the FCB feature. Similar to the center bias feature of image saliency detection \cite{judd2009learning}, we apply the following 2D Gaussian distribution to model $\mathbf{H}^f$ at each frame:
\begin{equation}
\label{Gauss_sigma}
\mathbf{H}^f(u,v)= \exp\left({-\frac{(u-u_f)^2+(v-v_f)^2}{\sigma_f^2}}\right),
\end{equation}
where $(u,v)$ are the longitude and latitude of the GCS location in the map, and $(u_f,v_f)$ are the longitude and latitude of the front center position in GCS. In addition, $\sigma_f$ is the standard deviation of  the 2D Gaussian distribution.

Next, we need to combine $\mathbf{H}^f$ with the predicted HM map $\mathbf{H}_t$ by
\begin{equation}
\label{optmization_x}
\mathbf{H}^c_t = w_1\cdot \mathbf{H}^f+w_2\cdot \mathbf{H}_t,
\end{equation}
for each panoramic frame. In the above equation, $\mathbf{H}^c_t$ is the HM map integrated with the FCB feature for frame $t$; $w_1$ and $w_2$ are the weights corresponding to the channels of $\mathbf{H}^f$ and $\mathbf{H}_t$, respectively. Given \eqref{Gauss_sigma} and \eqref{optmization_x}, the following optimization formulation is applied to obtain the values of $\sigma_f$, $w_1$ and $w_2$:
\begin{equation}
\label{optmization_w}
\max_{\sigma_f,w_1,w_2} \sum_{t=1}^{T} \text{CC}(\mathbf{H}^c_t, \mathbf{H}^g_t), \quad \text{s.t.} \quad w_1+w_2=1.
\end{equation}
In \eqref{optmization_x}, $\mathbf{H}^g_t$ is the ground-truth HM map of each frame; $\text{CC}(\cdot,\cdot)$ indicates the CC value of two maps. Then, we solve the above optimization formulation by the least square fitting of CC over all training data of our PVS-HM database. Consequently, the optimal values of $\sigma_f$, $w_1$ and $w_2$ are $21.1^\circ$, $0.48$ and $0.52$, respectively. These values are used to integrate the FCB feature in our offline-DHP approach. Note that the same way is applied to obtain the weights of $w_1$ and $w_2$, when combining the FCB feature with other approaches.

\begin{figure}
\vspace{-1em}
	\begin{center}
		\centerline{\includegraphics[width=.75\columnwidth]{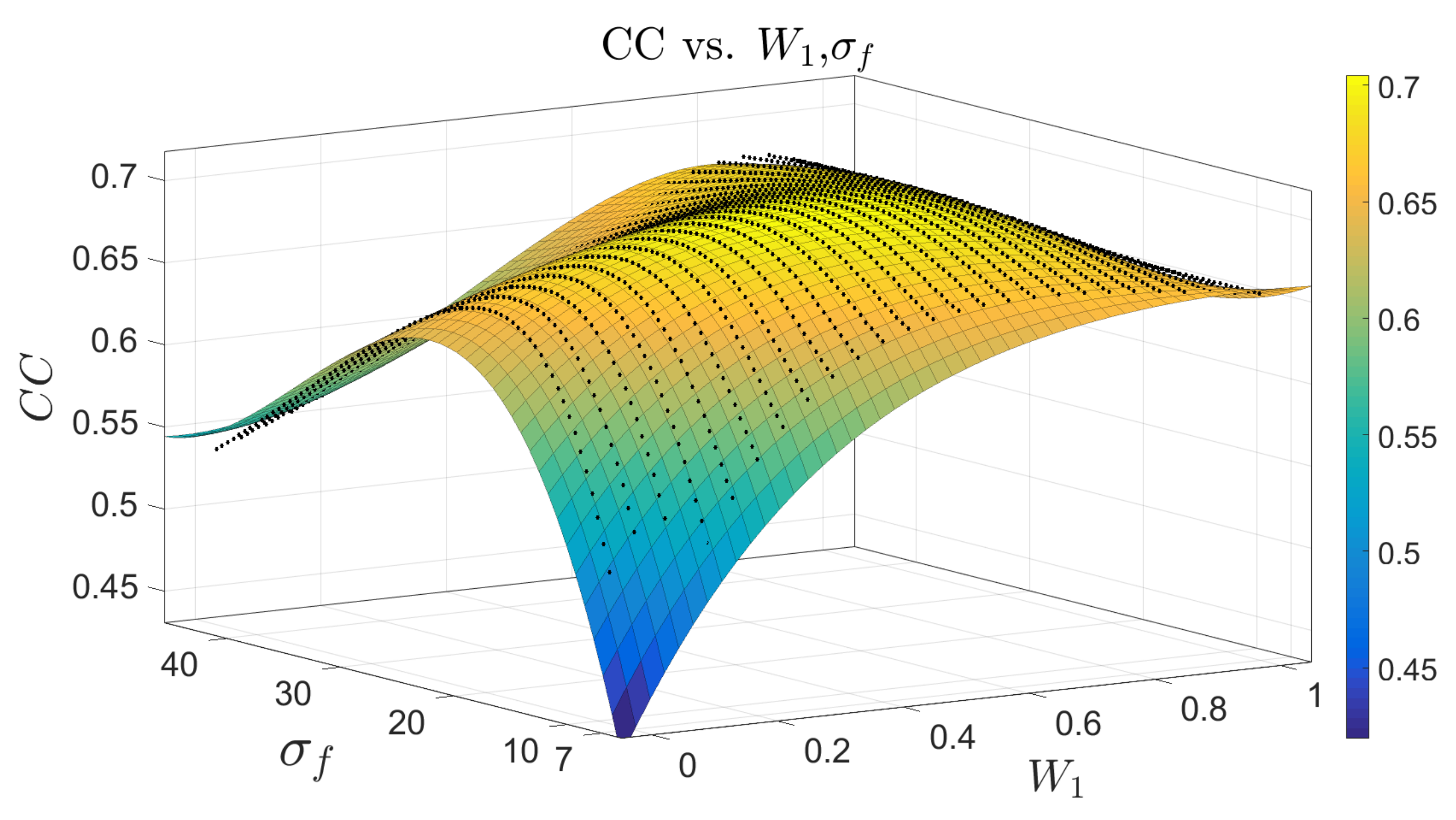}}
      \vspace{-1em}
		\caption{\footnotesize{The fitting surface of CC results between the predicted and ground-truth HM maps at various $\sigma_f$ and $w_1$. The dark dots in this figure represent the CC results at each specific value of $\sigma_f$ and $w_1$, which are used to fit the surface. Note that the CC results are obtained over all training data of the PVS-HM database.}}
		\label{fitting_surface}
	\end{center}
\vspace{-2.5em}
\end{figure}

Figure \ref{fitting_surface} shows the results of  CC between the predicted and ground-truth HM maps at various values of $\sigma_f$ and $w_1$. From this figure, we can see that the CC results vary from $0.44$ to $0.70$ alongside the increase of $w_1$ from $0$ to $1$, reaching the maximum value at $w_1=0.48$ given $\sigma_f=21.1^\circ$. This indicates that both the FCB feature and our offline-DHP approach are effective in predicting the HM maps of panoramic video, and that the effectiveness of the FCB feature is different at varying combination weights. In addition, as shown in Figure \ref{fitting_surface}, at $w_1=0.48$, the CC value increases from 0.66 to 0.70, when $\sigma_f$ grows from $7^\circ$ to $21.1^\circ$, and then it decreases to 0.63 until $\sigma_f = 43.6^\circ$. Thus, the standard deviation of the 2D Gaussian distribution in \eqref{Gauss_sigma} is set to be $21.1^\circ$ for the FCB feature in our experiments.

\ifCLASSOPTIONcompsoc
\else
\fi
\ifCLASSOPTIONcaptionsoff
  \newpage
\fi

\bibliographystyle{IEEEtran}
\bibliography{pami2017_dhp}

\begin{IEEEbiography}[{\includegraphics[width=1\linewidth]{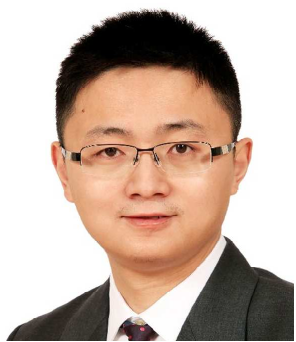}}]{Mai Xu}
(M'10, SM'16) received B.S. degree from Beihang University in 2003, M.S. degree from Tsinghua University in 2006 and Ph.D degree from Imperial College London in 2010. From 2010-2012, he was working as a research fellow at Electrical Engineering Department, Tsinghua University. Since Jan. 2013, he has been with Beihang University as an Associate Professor. During 2014 to 2015, he was a visiting researcher of MSRA. His research interests mainly include image processing and computer vision.  He has published more than 60 technical papers in international journals and conference proceedings, e.g., IEEE TIP, CVPR and ICCV. He is the recipient of best paper awards of two IEEE conferences.
\end{IEEEbiography}

\vspace{-2em}

\begin{IEEEbiography}[{\includegraphics[width=.80\linewidth]{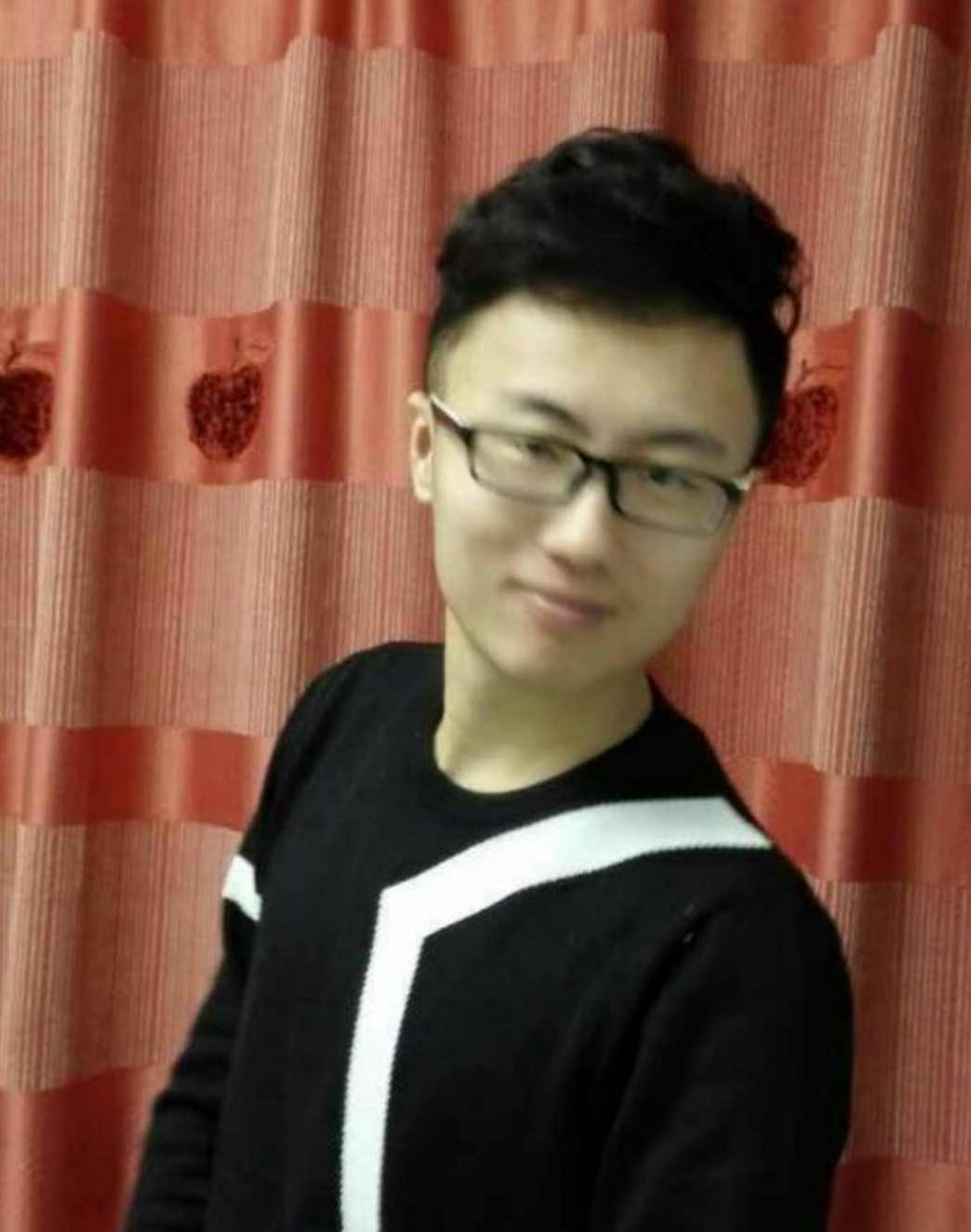}}]{Yuhang Song} is a research fellow from Beihang University. During 2017, he was a visiting researcher of Humanity-Centered Robotics Initiative at Brown University. In 2018, he will start his PhD study in University of Oxford. His research interests mainly include reinforcement learning and related applications. He has published several technical papers in the international conference proceedings, e.g., IEEE DCC.
\end{IEEEbiography}

\vspace{-2em}

\begin{IEEEbiography}[{\includegraphics[width=.90\linewidth]{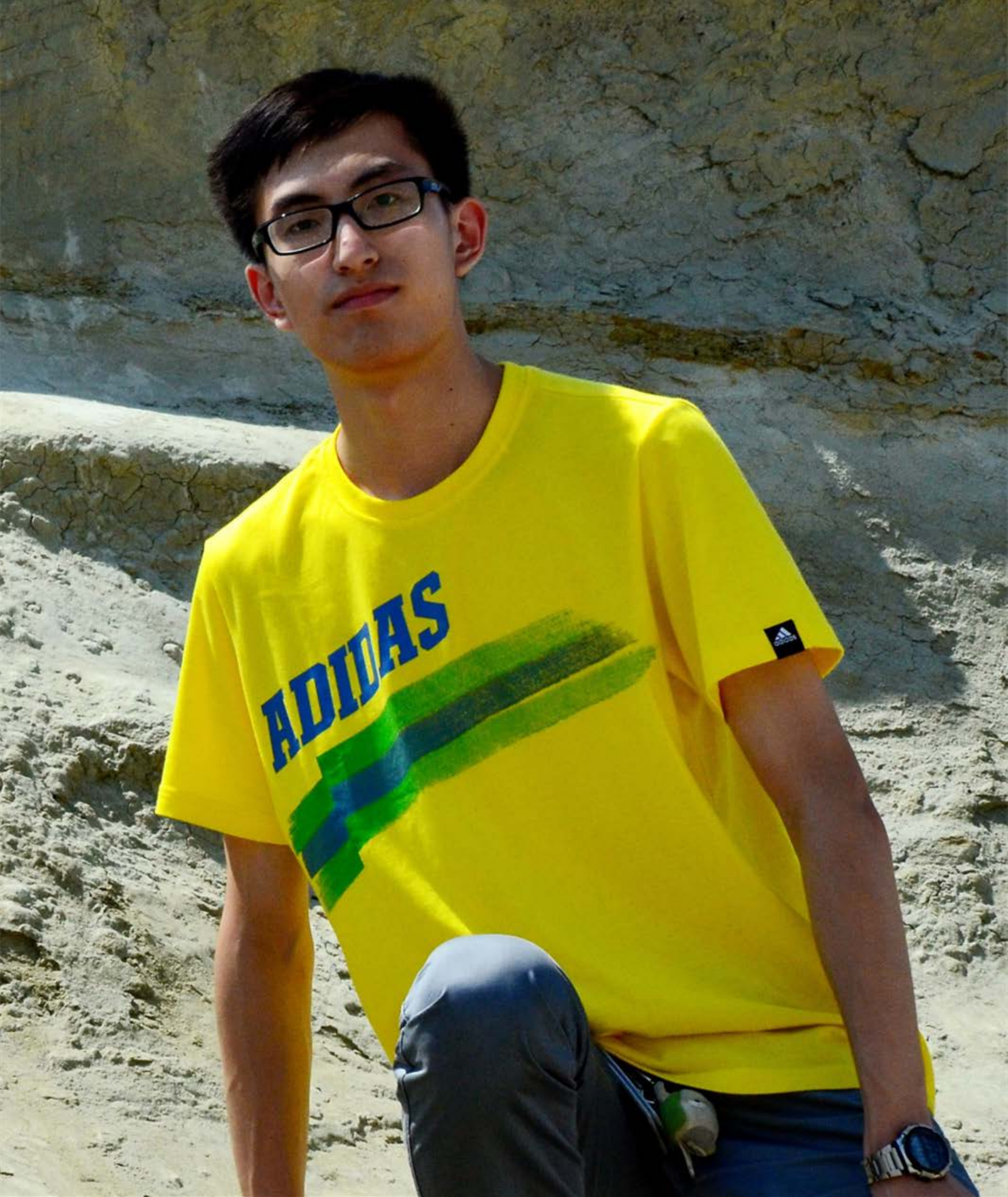}}]{Jianyi Wang} is a research assistant of Beihang University. He received his B.S. degree in the electrical enginneering from Beihang University in 2018 and is now a master student of Beihang University. He participated in lots of science and technology competitions and won many prizes during his undergraduate phase, e.g., 1st prize in "Challenge Cup" National Entrepreneurship Competition (The top entrepreneurship competition in China). Now his work is mainly related to panoramic video and reinforcement learning.
\end{IEEEbiography}

\vspace{-2em}

\begin{IEEEbiography}[{\includegraphics[width=.90\linewidth]{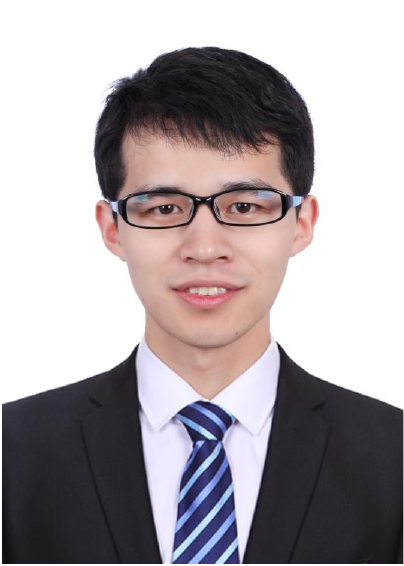}}]{Minglang Qiao} is a master student of Beihang University. He received the bachelor’s degree in the electrical enginneering from Beihang U- niversity, Beijing, China, in 2018. He is currently working towards the master’s degree at the MC2 Lab, Beihang University. His current research mainly focus on saliency detection of images and panoramic videos. He has published papers in the international conferences, e.g., ECCV.
\end{IEEEbiography}

\vspace{-2em}

\begin{IEEEbiography}[{\includegraphics[width=.90\linewidth]{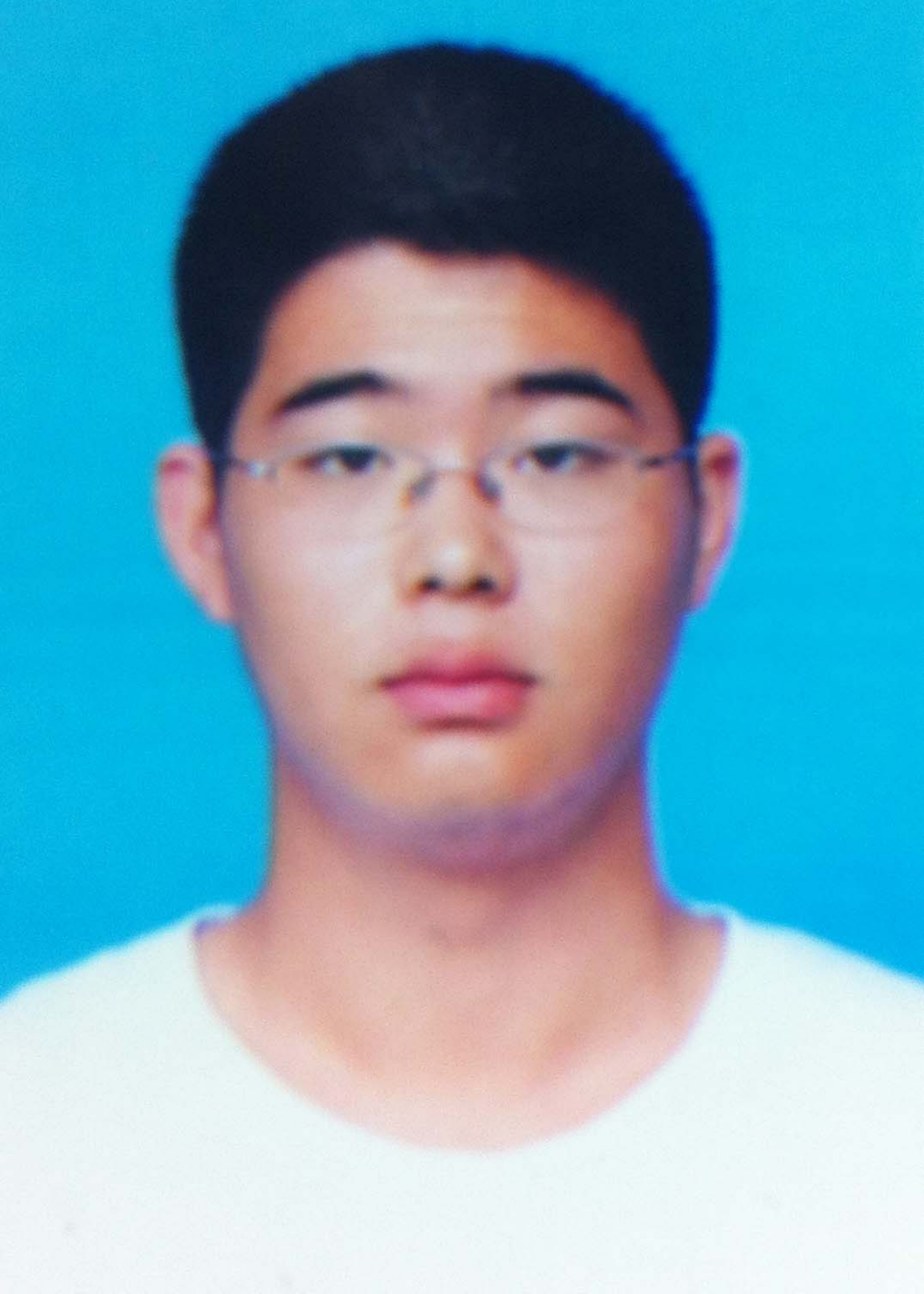}}]{Liangyu Huo} is a PhD candidate of Beihang University. He received his bachelor’s degree in the electrical enginneering from Beihang University, Beijing, China, in 2017. He participated in several science and technology competitions and won 1st prize in National College Student Information Security Contest. His research interests include multi-task reinforcement learning and hierarchical reinforcement learning.
\end{IEEEbiography}

\vspace{-10em}

\begin{IEEEbiography}[{\includegraphics[width=.80\linewidth]{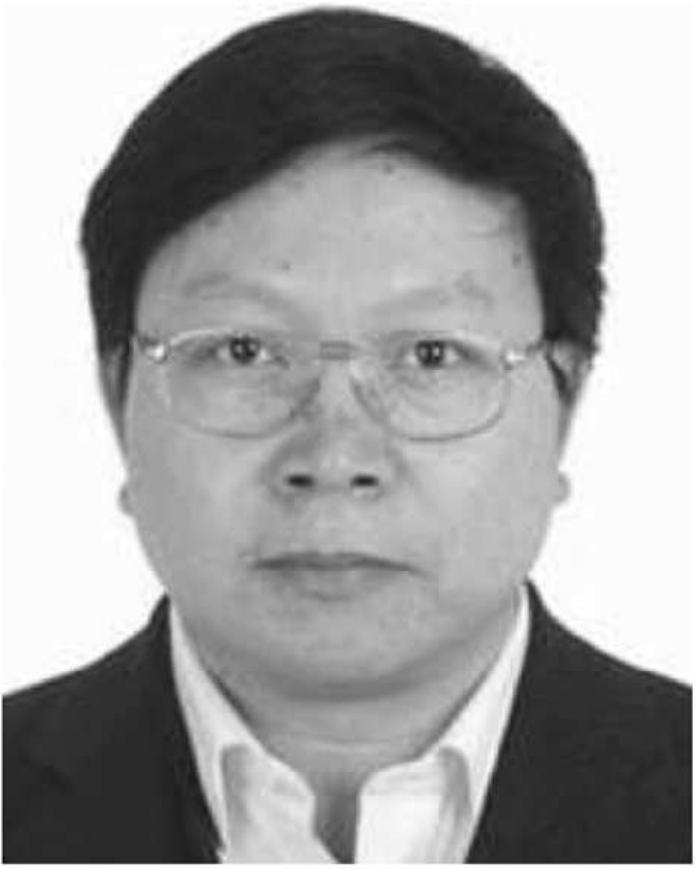}}]{Zulin~Wang}
received the B.S. and M.S. degrees in electronic engineering from Beihang
University, in 1986 and 1989, respectively. He received his Ph.D. degree at the same university
in 2000. He is currently a full professor at Beihang University, Beijing, China. He was the former dean of school of electronic and information engineering, Beihang
University. His research interests include image processing and remote sensing
technology. He is author or co-author of over 100 papers and holds 6 patents, as well as published 2 books in these fields. He has
undertaken approximately 30 projects related to image/video coding, image processing, etc.
\end{IEEEbiography}

\vspace{-2em}

\enlargethispage{-5in}

\end{document}